\documentclass[]{bmcart}
\pdfoutput=1 
\usepackage{booktabs} 
\usepackage{algorithm}
\usepackage{algorithmic}
\usepackage{subcaption}
\usepackage{url}
\usepackage{physics}
\usepackage{amsmath}
\usepackage{amssymb}
\usepackage{cleveref}
\DeclareMathOperator*{\argmin}{\arg\!\min}
\DeclareMathOperator*{\argmax}{\arg\!\max}
\usepackage{graphicx}
\usepackage{comment}

\begin{document}

\setcounter{page}{1}
\pagenumbering{arabic}
\begin{frontmatter}
\begin{fmbox}
\dochead{Research}

\title{Understanding the Limitations of Network Online Learning}

\author[
   addressref={aff1},
   corref={aff1},
   email={larock.t@husky.neu.edu}
]{\inits{TL}\fnm{Timothy} \snm{LaRock}}
\author[
   addressref={aff1},
   email={sakharov.t@husky.neu.edu}
]{\inits{TS}\fnm{Timothy} \snm{Sakharov}}
\author[
   addressref={aff3},
   email={sahely@iitpkd.ac.in}
]{\inits{SB}\fnm{Sahely} \snm{Bhadra}}
\author[
   addressref={aff1,aff2},
   email={tina@eliassi.org}
]{\inits{TER}\fnm{Tina} \snm{Eliassi-Rad}}


\address[id=aff1]{
  \orgname{Network Science Institute, Northeastern University},
  \city{Boston, MA},
  \cny{USA}
}

\address[id=aff2]{
  \orgname{Khoury College of Computer Sciences, Northeastern University},
  \city{Boston, MA},
  \cny{USA}
}

\address[id=aff3]{
  \orgname{Department of Computer Science, Indian Institute of Technology, Palakkad},
  \city{Palakkad, Kerala},
  \cny{India}
}


\begin{artnotes}
\end{artnotes}

\end{fmbox}

\begin{abstractbox}
\begin{abstract}
Studies of networked phenomena, such as interactions in online social media, often rely on incomplete data, either because these phenomena are partially observed, or because the data is too large or expensive to acquire all at once. 
Analysis of incomplete data leads to skewed or misleading results. 
In this paper, we investigate limitations of learning to complete partially observed networks via node querying. 
Concretely, we study the following problem: given (i) a partially observed network, (ii) the ability to query nodes for their connections (e.g., by accessing an API), and (iii) a budget on the number of such queries, sequentially learn which nodes to query in order to maximally increase observability. 
We call this querying process \textbf{N}etwork \textbf{O}nline \textbf{L}earning and present a family of algorithms called \textbf{NOL*}. 
These algorithms learn to choose which partially observed node to query next based on a parameterized model that is trained online through a process of exploration and exploitation. 
Extensive experiments on both synthetic and real world networks show that (i) it is possible to sequentially learn to choose which nodes are best to query in a network and (ii) some macroscopic properties of networks, such as the degree distribution and modular structure, impact the potential for learning and the optimal amount of random exploration.
\end{abstract}

\begin{keyword}
\kwd{Partially observed networks}
\kwd{online learning}
\kwd{heavy-tailed target distributions}
\end{keyword}
\end{abstractbox}
\end{frontmatter}


\newcommand{\VV}{\mathbb{V}}
\newcommand{\V}{\mathcal{V}}
\newcommand{\N}{\mathcal{N}}
\newcommand{\Ss}{\mathcal{S}}
\newcommand{\A}{\mathcal{A}}
\newcommand{\R}{\mathcal{R}}
\newcommand{\Rs}{\mathbf{R}}
\newcommand{\Ps}{\mathcal{P}}
\newcommand{\hide}[1]{}

\newcommand{\hG}{\hat{G}}
\newcommand{\hV}{\hat{V}}
\newcommand{\hE}{\hat{E}}

\newcommand{\note}[1]{{\color{red}{\bf #1}}} 
\newcommand{\todo}[1]{{\color{blue}{\bf TODO: #1}}}
\newcommand{\ignore}[1]{}
\renewcommand{\algorithmicrequire}{\textbf{Input:}}
\renewcommand{\algorithmicensure}{\textbf{Output:}}
\newcommand{\HTR}{NOL-HTR}
\newcommand{\NOLOR}{NOL}
\newcommand{\NOLSTAR}{NOL*}

\newcommand{\RR}{\mathbb{R}}

\section{Introduction}

\begin{figure}
	\includegraphics[width=0.99\textwidth]{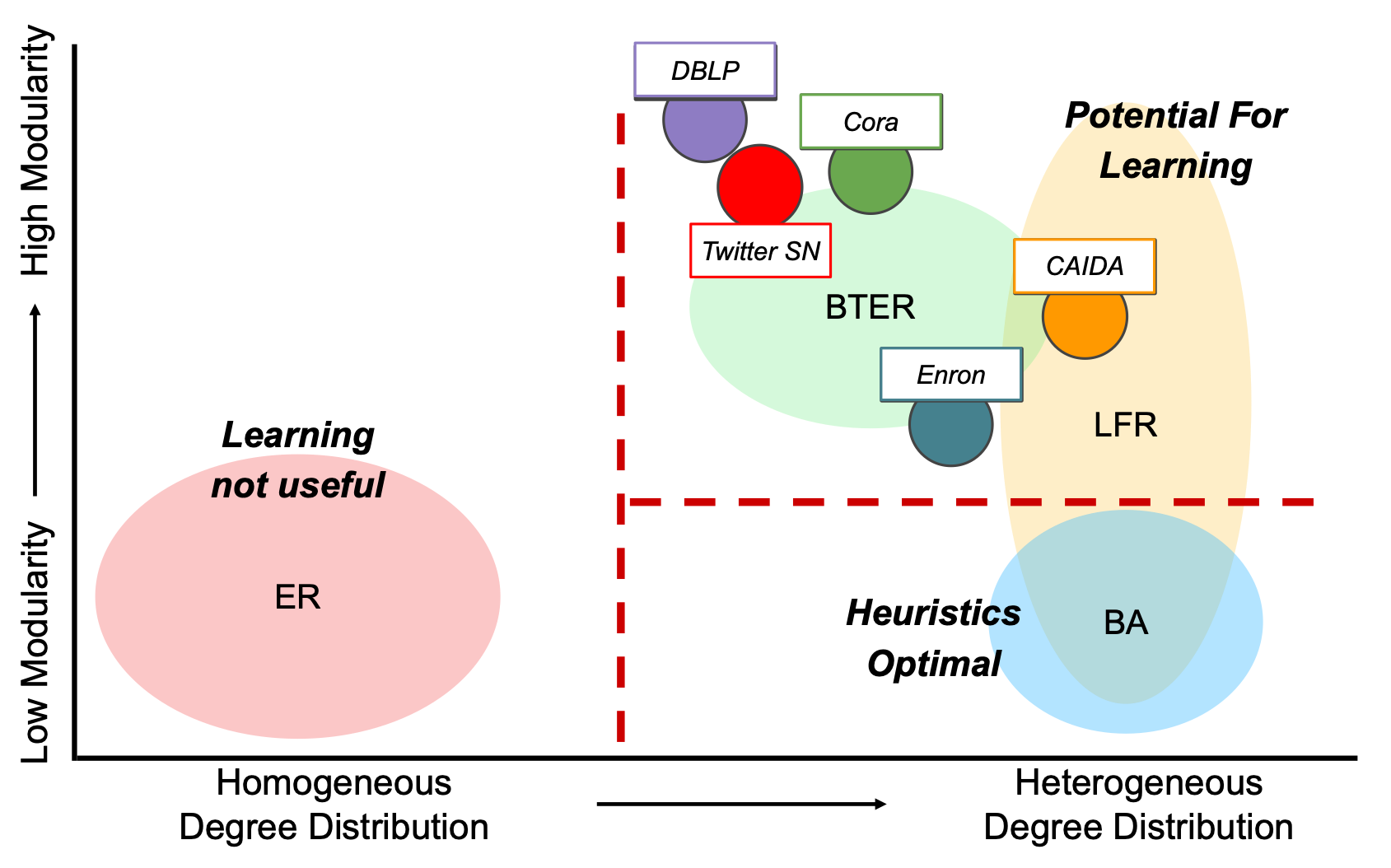}
	\captionsetup{width=0.9\textwidth}
	\caption{We present the conditions under which network online learning is feasible. In particular, there is a potential for learning a policy to reduce incompleteness of a network \emph{only  if}  the network has heterogenous degree distribution and high modularity. The synthetic graphs in the figure are ER (Erd\"{o}s-R\'{e}nyi), BA (Barab\'{a}si-Albert),  BTER (Block Two-level Erd\"{o}s-R\'{e}nyi), and  LFR (Lancichinetti-Fortunato- Radicchi).}
	\label{fig:takeaway}
\end{figure}

Incomplete datasets are common in the analysis of networks because the phenomena under study are often partially observed. 
%
%
 It has been shown that analysis of incomplete networks may lead to biased results \cite{Ghosh2013, Moreno2014, Sampson2015}. Our work seeks to address the following problem: \textit{Given a partially observed network with no information about how it was observed and a budget to query the partially observed nodes, can we learn to sequentially ask optimal queries relative to some objective? } In addition to introducing new methodology, we study when learning is feasible in this problem (see \cref{fig:takeaway}).


We present a family of online learning algorithms, called \NOLSTAR, which are based on online regression and related to reinforcement learning techniques like approximate Q-learning. \NOLSTAR\ algorithms do not assume any \emph{a priori} knowledge or estimate of the true underlying network structure, overall size, or the sampling method used to collect the partially observed network.  

We describe two algorithms from the \NOLSTAR\ family, both of which learn an \emph{interpretable} policy for growing an incomplete network through successive node queries.  Interpretability of these algorithms is important because we want to be able to reason about when and why a policy is or is not learnable.  The first algorithm, referred to as \NOLOR\footnote{\footnotesize An earlier version of \NOLOR\ was presented at the 14th Annual Workshop on Mining and Learning with Graphs, a non-archival venue, in 2018 \cite{LaRock2018}.}, uses online linear regression as a model for predicting the value of querying available nodes, then uses those predictions to choose the best node to query next. The second algorithm, \HTR, uses a heavy-tail regression method to account for heavy-tailed (equivalently heterogeneous) reward functions. This is necessary because in the case of a heavy-tailed reward distribution such as node degree, \NOLOR\ will under-predict the objective for hubs (which are the ``big and rare'' instances). 

We conduct experiments using \NOLSTAR\ algorithms on graphs generated by synthetic models, as well as real-world network data.\footnote{\footnotesize We use the terms graph and network interchangeably in this paper.} We focus on the following objective or reward function: discover the maximum number of initially unobserved nodes in the network through successive node querying. 

\paragraph{Formal problem definition}
Given an \emph{incomplete} network $\hG_0 = \{\hV_0, \hE_0\}$, which is a partial observation of an underlying network $G = \{V,E\}$, sequentially learn a policy that maximizes the number of nodes $u \in \hV_b, u \notin \hV_0$ after $b$ \emph{queries} of the incomplete network. Probing the network involves selecting a node and asking an oracle or an API for all the neighbors of the selected node. 

This problem is a sequential decision-learning task that can be formulated as a Markov Decision Process (MDP), where the \emph{state} of the process at any time step is the partially observed network, the \emph{action space} is the set of partially observed nodes available for probing, and the \emph{reward} is a user-defined function (e.g., the increase in the number of observed nodes). The goal of an MDP learning algorithm is to learn a mapping from states to actions such that an agent using this mapping will maximize its expected reward. In our case, the state-action space of the problem can be arbitrarily large given that we make no assumptions about the underlying network; thus any solution will need to learn to generalize over states and actions from experience (i.e., answers to successive queries).  

Our framework \NOLSTAR\ can be viewed as solving an MDP problem. Its algorithms learn models to predict the expected reward gained by probing a partially observed node. The current model is then used at each time step to decide the best action (i.e., which node to query to observe as many new nodes as possible). In this way, \NOLSTAR\ algorithms choose the action that leads to maximizing the total reward over time within an arbitrary budget constraint.

\NOLSTAR\ algorithms query nodes in the partially observed network \emph{sequentially}.  That is, in each iteration the algorithm queries one node, adds all of its neighbors to the observed network, and updates the pool of partially observed nodes available to be queried in the next iteration. This makes \NOLSTAR\ algorithms \emph{adaptive}, since the parameters of the model change based on recent experience. Initially, all nodes in the network are assumed to be partially observed, thus \NOLSTAR\ is agnostic to the underlying observation or sampling method. At any iteration, there are three ``classes'' of nodes: fully observed (probed), partially observed (unprobed but visible) and unobserved (unprobed and invisible).\footnote{\footnotesize Here we assume that the underlying network is static.  However, our approach can be extended to allow for repeated probing of nodes under any API access model, which is useful when the underlying network is not static.}

\paragraph{Contributions} Our contributions are as follows:
\begin{itemize}
\item We propose a family of interpretable algorithms, \NOLSTAR, for learning to grow incomplete networks.
\item We present two algorithms from the \NOLSTAR\ family: an online regression algorithm, simply called \NOLOR; and an algorithm that can effectively learn for heavy-tailed objective functions, called \HTR.
\item Our extensive experiments on both synthetic and real network data showcase the limitations of online learning to improve incomplete networks.
\end{itemize}

In the next section we summarize the literature related to the network discovery problem. Then, we describe and evaluate \NOLSTAR\ before closing the paper with a discussion of future directions.
\section{Related Work}
\label{sec:relatedwork}
Incomplete data sets affect numerous areas of research from social network analysis of public health \cite{Gile2011,Wejnert2008} and economics \cite{Breza2017}, to mining of the World Wide Web (WWW) \cite{Cho2012,Avrachenkov2014} and analyzing Internet infrastructure  \cite{Vazquez2002}. 

The problem of network online learning is different from network sampling. In traditional network sampling, the goal is to gather a representative sample of the underlying network from which statistical characteristics are then estimated. In our setting, the data collection is guided by an initial sample graph and a user-defined objective function, which may or may not be directly related to a notion of statistical representativeness of the data.
For an excellent survey on network sampling, we refer the reader to~\cite{Ahmed2014}. 

One approach to growing incomplete networks is to assume that the graph is being generated by an underlying network model, then use that model to infer the missing nodes and links.  Examples of this approach are ~\cite{Hanneke2009} and~\cite{Myunghwan2011}.  The former assumes stochastic block models~\cite{sbm}, while the latter assumes Kronecker graph models~\cite{Leskovec2010}. We refer the reader to~\cite{modelselection} for a recent paper on model selection for mechanistic network models.

Avrachenkov et al.~\cite{Avrachenkov2014} explore methods for maximally covering a graph by adaptive crawling and querying. Their work introduces the Maximum Expected Uncovered Degree (MEUD) method and shows that for a certain class of random power law networks, MEUD reduces to choosing the node with maximum observed degree for each query (which is equivalent to our High Degree heuristic; see Section~\ref{sec:exp}).

MAXOUTPROBE~\cite{Soundarajan2015} estimates the degree of each observed node and the average clustering coefficient of the graph; it does not assume knowledge of how the incomplete network was collected.  MAXREACH~\cite{Soundarajan2016} estimates the degree of each observed node and the per-degree average clustering coefficient.  It assumes that the incomplete network was collected via random node or edge sampling; and assumes that one knows the size of the fully observed graph in terms of its number of nodes and edges.

Multi-armed bandit approaches are well-suited for the problem of growing incomplete networks because they are designed explicitly to facilitate the exploration vs.~exploitation tradeoff. Soundarajan et al.~\cite{Soundarajan2017} present a multi-armed bandit approach to network completion that trades off densification vs.~expansion.  Their approach also focuses on the problem of probing edges rather than nodes.  Murai et al.~\cite{Murai2018} propose a multi-armed bandit algorithm for reducing network incompleteness that probabilistically chooses from an ensemble of classifiers that are trained simultaneously.  Madhawa and Murata~\cite{madhawa2019multiarmed} describe a nonparametric multi-armed bandit based on a k-nearest neighbor upper-confidence bound algorithm, which will be described in more detail in \cref{sec:exp}.



Active Exploration~\cite{Pfeiffer2014} and Selective Harvesting~\cite{Murai2018} address variants of the problem definition where the general task is to iteratively \emph{search} a partially observed network through node querying. The goal is to maximize the number of nodes in the expanded network with a particular binary target attribute. We address a different problem, where the goal is to maximally grow the network, in this paper.

\NOLSTAR\ provides a unified framework for interpretable and scalable network online learning. It incorporates exploration vs.~exploration.  It does not assume how the incomplete network was originally collected or a specific model generating the underlying network.

\section{\NOLSTAR\ Family of Algorithms}
\label{sec:model}
Algorithm \ref{alg:nol} presents the general \NOLSTAR\ framework. The goal is to sequentially learn to predict the reward value of partially observed nodes in an incomplete network under a resource constraint or budget, $b$, on the number of queries we can ask an oracle or API, leveraging previous queries as sample data. One node is queried at every time step ${t = 0, 1, 2, \ldots b}$. The partially observed network $\hat{G}_t =\{\hV_t,\hE_t\} \subset G$, with nodes $V_t$, edges $E_t$ and the list of nodes which have been already probed $P_t$, are updated by the algorithm after every query.  The incomplete network $\hG_t$ grows after every probe by incorporating the neighbors of the probed node $j$, selected from $(\hat{V}_t - P_t)$. The number of new nodes added to the observed network by probing $j$ in timestep $t$ is the \emph{reward} earned in that timestep, denoted by $r_t(j)$.

The goal is to make a prediction about the reward to be earned by querying any partially observed node $j$ that maximizes reward at time $t$. In general, given an initial state $s_0$, we wish to maximize the \emph{cumulative reward} earned after $b$ queries, given that the starting point was $s_0$: 
$$c_r(b) = \sum_{t=0}^{b}r_t|s_0$$

In each time step $t$, we want to maximize the reward $r_t$ earned by taking action $a$ from state $s_t$. Therefore we learn to choose an action, which corresponds to choosing a node $u$ to query, such that 
$$u=\argmax_{j\in(\hat{V}_t - P_t)} r_t(j)$$ 
where $r_t(j)$ is the reward earned by querying node $j$ from state $s_t$. 

Although a more general Temporal Difference Learning \cite{sutton2018reinforcement} solution to this problem can be formulated, we did not find an advantage in using discounting or credit assignment in experiments. We conjecture that this is due to a combination of the finite resource constraint and the fact that we do not visit the same states multiple times in our setting, making standard planning tools less effective.

\begin{algorithm}
\footnotesize
\caption{\NOLSTAR\ Framework}
\begin{algorithmic}[1]
\REQUIRE $\hG_0$ (initial incomplete network), $b$ (probing budget),  $\epsilon_0$ (initial jump rate), $k$ and $\lambda$ (for \HTR), $\alpha$ (for \NOLOR).
\ENSURE $\hG_b$ (network after $b$ probes), $\theta_b$ (learned parameters).
	\STATE Initialize: $\theta_0$ (randomly or heuristically); $P_0, X, Y = \emptyset$; $t=0$; $\epsilon_t = \epsilon_0$
	\REPEAT 
		\STATE  \textrm{Calculate feature vectors:} $\phi_t(i)$, $ \forall \textrm{ nodes } i \in \hV_t-P_t$ 
		\STATE  \textrm{Calculate estimated rewards:} $\V_{\theta_t}(i) =\theta_{t}\phi_t(i)$ 
		\STATE \textrm{Explore: With probability} $\epsilon_t$, choose $u_t$ uniformly at random from $\hV_0-P_t$.
		\STATE \textrm{Exploit: With probability} $1-\epsilon_t$, \\ $u_t = \argmax_{i \in \hV_t-P_t} \V_{\theta_t}(i)$
		\STATE Probe node $u_t$, add to $P_{t+1} = P_{t}\cup u_t$
		\STATE Update the observed graph: $\hG_{t+1} = \{\hG_t \cup \textrm{ neighbors of }u_t\}$ 
		\STATE Collect reward: $r_t(u_t)= |\hV_{t+1}| - |\hV_t|$
	    \IF{\NOLOR}
			\STATE $loss_t=  \left( r_t - \V_{\theta_t}(u_t)\right)^2$
			\STATE $\nabla_{\theta_t}loss_t = -2 \left(r_t - \V_{\theta_t}(u_t) \right)  \phi_t(u_t)$ 
			\STATE $\theta_{t+1}= \theta_t - \alpha  \nabla_{\theta_t}loss_t$
		\ELSIF{\HTR} 
			\STATE Append $\phi_t(u_t)$ to $X$ and $r_t$ to $Y$
			\STATE $\theta_{t+1} = HeavyTailRegression(X, Y, k, \lambda)$
		\ENDIF
		\STATE $t = t+1$
		\IF{decay} \STATE $\epsilon_{t}=\epsilon_0 e^{\frac{-t}{b}}$
		\ENDIF
	\UNTIL {t==b}
	\RETURN $\theta_b$ and $\hG_b$ 
\end{algorithmic}
\label{alg:nol}
\end{algorithm}

\begin{algorithm}
\footnotesize
\caption{$HeavyTailRegression$}
\begin{algorithmic}[1]
\REQUIRE $\mathbf{X}$ (matrix of probed node features),  $Y$ (associated reward values), $k$, and $\lambda$
\ENSURE $\theta$ (updated parameters)
	\STATE Partition data into k uniformly random subsamples $S_1, \dots S_k$ of size $\frac{n}{k}$, where each $ S_i = \mathbf{X_i}\subset \mathbf{X}, Y_i \subset Y $
	\FOR{each subsample $S_i$} 
		\STATE Compute covariance $\Sigma_i$ of $\mathbf{X_i}$
        \STATE Compute regression parameters $\omega_i \in \argmin_{\omega}L^\lambda(\mathbf{X_i}, Y_i)$ 
	\ENDFOR
    \FOR{each subsample $S_i$} 
        \STATE Compute $m_i = \textbf{median}_{j\neq i} \langle(\omega_i-\omega_j), (\Sigma_{S_j}+\lambda Id)(\omega_i+\omega_j)\rangle$
    \ENDFOR
    \STATE Assign $\theta$ to be the $\omega_i$ associated with the minimum $m_i$
	\RETURN $\theta$ 
\end{algorithmic}
\label{alg:htr}
\end{algorithm}

To predict the reward value of every node in $\hG_t$, a feature vector $\phi_t(j) \in \RR^d$ is maintained that represents the knowledge available to the model at time $t$ for node $j \in \hat{G}_t$. Given features $\phi_t(i)$ for all nodes in $\hG_t$, \NOLSTAR\ algorithms learn a function $\V_{\theta}: \RR^d \rightarrow \RR$ with parameter $\theta$ to predict the expected reward to be earned by probing the node. Any available information about a node can be included as a feature, but since learning is to happen online, features should be feasible for online computation. Further, it is desirable for features to be interpretable so that the resulting regression parameters can be interpreted to help understand the performance of the algorithm in a given dataset (see \cref{fig:app-feature-weights}).

Given parameter $\theta_t$ at time $t$, the predicted number of unobserved nodes attached to node $j \in \hV_t$ is estimated as $\V_{\theta_t}(j) = \theta_t\phi_t(j)$ such that $\V_{\theta_t}(j)$ should be close to the observed reward value of probing the node $j$ at time $t$. At each step, the expected loss $E_t[\V_{\theta_t}(j) - r_t(j)]$ is minimized, where $r_t(j)$ is the true value of the reward function for node $j$ at time $t$. 

In \NOLOR, $\theta$ is updated after each probe through online stochastic gradient descent based on \cite{strehl2008online} (see lines 10-13 of Algorithm \ref{alg:nol}). 
If the variable of interest is heavy-tailed (e.g., the degree distributions of real networks exhibit this property), \HTR\ adopts the generalized median of means approach to regression with heavy tails found in \cite{Hsu2016}. These example functions illustrate the flexibility of \NOLSTAR: the choice of reward function and learning algorithm should correspond to individual goals and circumstances. We describe the details of how we learn the parameters for \HTR\ in the next section and \cref{alg:htr}.

\subsection{Parameter Estimation for \HTR}

Our goal at every time $t$ is to choose a node that maximizes the earned reward. However, since the reward distribution is based on node degree, and the degree distribution in many real-world networks is heavy-tailed (e.g. hubs are present), then we expect the reward distribution of an incomplete version of the network to be heavy-tailed as well. In order to deal with the heavy tailed nature of the target variable, we adapt the methods from \cite{Hsu2016} for regression in the presence of heavy-tailed distributions. This process is a generalization of the median of means approach to parameter estimation. Intuitively, the algorithm splits the previously observed feature vectors and associated rewards into $k$ subsamples, then computes parameters $\omega$ for each subsample. Then, the algorithm chooses the set of parameters that has the minimum median distance from all of the other parameters. This procedure provides guarantees and confidence bounds on the distance of the learned parameters from the true parameters \cite{Hsu2016}.

Algorithm~\ref{alg:htr} presents our adopted process. At time $t$, $t-1$ nodes with feature vectors $x_i \in X$ have been probed, and their corresponding reward values $r_i\in Y$ observed. We select an integer $k \leq t$ and randomly sample the data into $k$ subsamples  $S_i$ of size $\frac{t}{k}$. Then, the covariance matrix $\Sigma_i$ and maximum likelihood regression estimate $\omega_i$ are computed for each $S_i$. For each $i$, the Mahalanobis distance between $\omega_i$ and every other $\omega_j$ is computed, and the median distance is assigned to $m_i$. Finally, the $\omega_i$ with the minimum $m_i$ value is assigned to be the next set of parameters, $\theta$.  

There are two parameters in Algorithm \ref{alg:htr}: the number of subsamples $k$, which corresponds to the confidence parameter $\delta$ in \cite{Hsu2016}, and the regularization parameter $\lambda$. The number of subsamples $k$ should be set such that the size of the subsamples, $\frac{n}{k}$, is larger than $O(d\log(d))$, meaning each subsample has size at least the number of dimensions in the feature vector \cite{Hsu2016}. In our experiments, we set $k$ to be $\ln{(n)}$, where $n$ is the number of previously queried nodes (equivalently the time step $t$), which allows $k$ to grow slowly as more data is gathered through the querying process. We set the regularization parameter $\lambda$ to 0. However, our feature matrices may be singular (since some subsamplings can result in feature matrices without full rank), so when computing the regression in Algorithm \ref{alg:htr} we use the Moore-Penrose pseudoinverse, which corresponds to computing the $\ell_2$ regularized parameters.

\subsection{$\epsilon$-greedy Exploration}
\NOLSTAR\ algorithms need to learn adaptively over time because the reward distribution may change as nodes are queried. Choosing a node based on $V_{\theta_t}$ \emph{exploits} the current model by choosing the node with the maximum predicted reward. However, learning in networks is difficult precisely because the properties of nodes are diverse, thus it is desirable to introduce some randomness to the decision process in order to gather a diverse set of training examples. Therefore, we formulate \NOLSTAR\ algorithms as $\epsilon$-greedy algorithms, meaning with probability $\epsilon$ the node to query is chosen uniformly at random from the set of unprobed nodes.

In order to increase the likelihood that our random samples are informative, we choose our random nodes from those that were present in the initial network $\hG_0$. The rationale behind this choice is that as a consequence of probing nodes sequentially nodes that have been in the network for longer have more ``complete" information, since they have had more opportunity to be connected to in the $t-1$ probes before time $t$. That is, if a node $j \in \hG_0$ has only a few neighbors after many queries have been made, it could be that $j$ has very few neighbors, but it could also be that $j$ connects to a neighborhood that the algorithm has yet to discover. Therefore, to explore the possibility of learning a better model using different information,  \NOLSTAR\ algorithms select the random node from $\hV_0 - P_t$ to probe until all nodes in $\hV_0$ are exhausted, when \NOLSTAR\ chooses any unprobed node in $\hG_t$ at random. 

Since \NOLSTAR\ algorithms learn from all or most of the previous samples at every time step, it is not strictly necessary to continue random exploration through the entire querying process. This is consistent with the literature on $\epsilon$-greedy algorithms, which often systematically lower the rate of exploration over time \cite{Kirkpatrick1983, Cheng2010}. \NOLSTAR\ adds an optional exponential decay to the initial random jump rate $\epsilon_0$, such that at time $t$ the jump rate is computed as $\epsilon_t = \epsilon_0 e^{\frac{-t}{b}}$. We have also experimented with data-driven methods for adaptive-$\epsilon$-greedy, such as in \cite{Tokic2010}, but did not find them to be advantageous and leave further investigation of their utility in this space for future work.

\subsection{Scalability \& Guarantees}
In general, the computational complexity of a \NOLSTAR\ algorithm is the product of (1) the budget, (2) the feature computation complexity and (3) the learning complexity. 
In symbols, $O(b\times O(\textrm{features}) \times O(\textrm{learning}))$.

In this section, we briefly discuss the complexity of the learning steps of \NOLOR\ and \HTR.
We also note that the features used in \NOLSTAR\ algorithms are user defined and range from trivial to compute (e.g. degree) to computationally expensive (e.g. node embeddings).
Due to this, we omit a detailed analysis of the specific features we use in our experiments (described in \cref{sec:exp}), but note that since the computations are to happen online, and only nodes whose features may have changed should be updated, the complexity depends not necessarily on the total number of nodes or edges in the graph, but on the size of the neighborhood around the queried node for which the features might have changed.

The complexity of a \NOLOR\ online regression update depends only on the number of features, since the most expensive operation required is multiplication of the feature vector by a constant factor. Due to this, even with a relatively high-dimensional feature vector, the complexity of the learning step is trivial. In this case, the scalability of the entire process will likely hinge on the scalability of the feature value updates.

The procedure to learn parameters for \HTR\ is more complicated. The data is first partitioned into k subsamples, the covariance of each subsample is computed and optimal regression parameters are estimated for each, and finally the median of the $k$ parameters is computed. The most expensive operation is the regression parameter estimation, which requires computing the Moore-Penrose inverse of the regressor matrix. This can be computed via Singular Value Decomposition of the matrix, which has complexity $O(nd^2)$, where $n$ is the number of nodes in the computation and and $d$ is the number of features. 

Hsu and Sabato \cite{Hsu2016} also derive a bound on the loss and show that our learned parameters $\theta$ are within an $\epsilon$ of the true parameters if we use \cref{alg:htr} with the number of samples $m\geq O(d\log(\frac{1}{\delta}))$. In our case, $m$ is the number of rows in $X$. The derived bound states that with probability $(1-\delta)$ the empirical loss is bounded by
$$L(\theta) \leq \left( 1+ O\left( \frac{d\log{(\frac{1}{\delta})}}{m}\right)\right)L_*$$ 
where $L_*$ is the true loss with the optimal parameters and $\delta=\frac{1}{e^k}$. In practice, we avoid wasted computation by limiting the total number of samples $m$ to 2000, which is always larger than is necessary for the guarantees given the values of $k$ in our experiments (which determines delta in our formulation of the algorithm).
\section{Experiments}
\label{sec:exp}

\begin{figure*}[!ht]
\centering
\includegraphics[width=0.95\columnwidth]{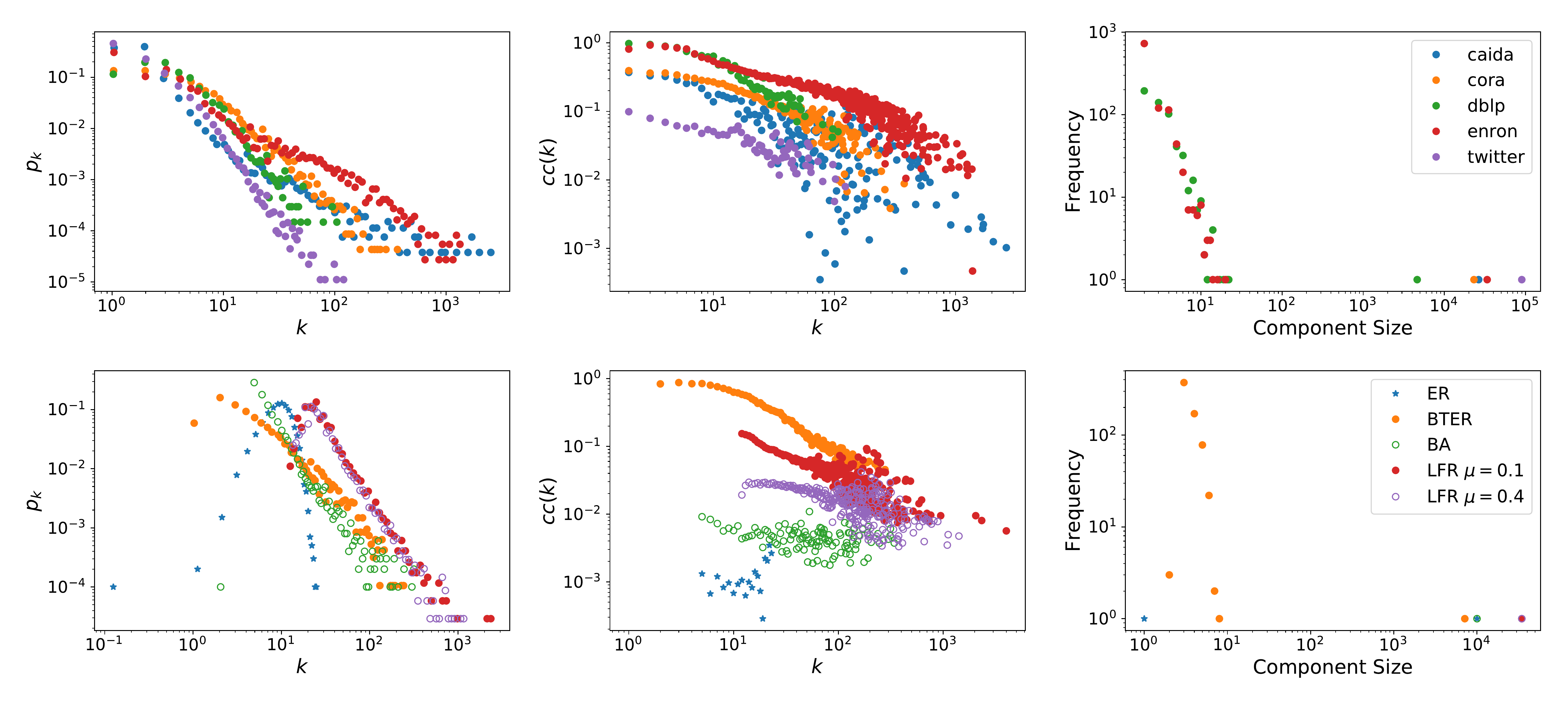}
\vspace{-0.3cm}
\captionsetup{width=0.9\textwidth}
\caption{The top row from left to right shows the (node) degree distribution, average (node) clustering coefficient per degree, and frequency of connected component sizes of five real-world networks. The bottom row shows the same quantities, but for three synthetic graph models: Erd\"{o}s-R\'{e}nyi (ER)~\cite{er}, Barab\'{a}si-Albert (BA)~\cite{ba}, and Block Two-Level Erd\"{o}s-R\'{e}nyi (BTER)~\cite{bter}. \HTR\ is able to learn to increase network size on BTER and similar real-world networks. We find that degree, clustering coefficient, and size of the connected component are all relevant, as well as interpretable, features for learning. 
}
\label{fig:distributions}
\end{figure*}

In this section, we show the utility of applying \NOLSTAR\ algorithms for network completion in a variety of datasets, including both synthetic and real world networks. We first describe the data we use to test our method, then explain our experimental methodology and present results.

\subsection{Data}
We test \NOLSTAR\ algorithms on a range of synthetic graphs, as well as five real world datasets. 

\subsubsection{Synthetic Models}
\label{sec:synthetic}
Synthetic network models are useful for generating test data that exhibits interesting properties often found in real world networks. We are interested in testing the performance of our algorithm on datasets that allow us to vary a few macroscopic properties of networks: the distribution of degrees, the extent of local clustering, the modularity of the global structure. 

The degree distribution is an important factor in understanding when learning is possible or helpful. There are two simplified extremes that can characterize degree distributions in complex networks: \emph{homogenous} and \emph{heterogenous} (or heavy tailed) distributions. The important difference is that in a heterogeneous degree distribution some nodes accumulate far more connections than the majority of nodes in the network, resulting in these nodes becoming topological hubs. In the presence of hubs, the variance of the degree distribution can become large as the number of nodes in the network $N$ grows, until it eventually diverges as $N\rightarrow\infty$. In contrast, a homogeneous degree distribution implies that hubs are not present, meaning the nodes are statistically equivalent with respect to degree. In the homogenous case, the degree of a randomly chosen node will be a random variable following a distribution with well-defined variance. 

A second node characteristic that we conjecture is important for learning is the \emph{clustering coefficient}. The clustering coefficient (more precisely, \emph{local} clustering coefficient) is a measure of the extent to which a node's neighbors are connected to one another. Since the clustering coefficient is real-valued, it is often more intuitive to study the average clustering by degree, which is what we show in Figure \ref{fig:distributions}. Clustering is related to the local density of neighborhoods and can serve as a proxy for the amount a neighborhood has been explored. 

Lastly, we are interested in studying the impact of the \emph{modularity} of a network structure on algorithm performance, meaning the prevalence of within-community links relative to out-of-community links in a modularity-maximized partitioning of the nodes into communities. The extent of modularity in a network structure could be instructive in balancing exploration and exploitation, since learning to probe in a highly modular structure is more susceptible to settling on local minima by exploiting in a single community without discovering cross-community links.

To test the above conjectures on the limitations of learning in complex networks, we study five synthetic network models. 
\begin{enumerate}
    \item  Erd\H{o}s-R\'{e}nyi\ (ER)~\cite{er}:\footnote{\footnotesize We omit almost all results on ER graphs from the paper because all methods perform indistinguishably on these graphs.} In a network sampled from the the ER model (specifically the ensemble $G_{Np}$), every possible (undirected) link between $N$ nodes exists with probability $p$. Networks generated by the ER model have a homogeneous degree distribution (the exact distribution is Binomial, but it is often approximated by Poisson). \underline{Parameters}: $N = 10000$, $p = 0.001$. 
    \item Barab\'{a}si-Albert (BA)~\cite{ba}: The BA model generates networks through a growth and preferential attachment process where each node entering the network chooses a set of $m$ neighbors to connect to with probability proportional to their relative degree. This process results in a heterogeneous degree distribution, which in the infinite limit follows a power law distribution with exponent 3. \underline{Parameters}: $ N = 10000$, $m = 5$, $m_0 = 5$. $m_0$ denotes the  the size of the initial connected network. $m$ denotes the number of existing nodes to which a new node connects.
    \item Block Two-level Erd\H{o}s-R\'{e}nyi (BTER)~\cite{bter}: BTER is a flexible\footnote{\footnotesize There are many other random graph models that provide flexibility similar to the BTER that we could have used to generate networks for our experiments. We chose BTER out of convenience because it allows us to easily specify target values for average degree and clustering parameters directly.} model that combines properties of the ER and BA model. It consists of two phases: (\emph{i}) construct a set of disconnected \emph{communities} made up of dense ER networks, with the size distribution of the communities following a heavy tailed distribution (i.e., a small number of large communities and many more small communities) and (\emph{ii}) connect the communities to one another to achieve desired properties, such as a target value of global clustering coefficient. \underline{Parameters}: $N = 10000$, target maximum clustering coefficient = 0.95, target global clustering coefficient = 0.15, target average degree $\langle k \rangle = 10$. 
    \item Lancichinetti-Fortunato-Radicchi (LFR)  \cite{LFR}: In the LFR model, modular structure can be induced by varying the mixing parameter $\mu$, which controls the extent to which nodes connect internally to a tight community (higher modularity) or loosely to the entire network (lower modularity), thus controlling the extent of modular community structure. \underline{Parameters}: $N = 34546$, mixing parameter $mu \in 0, 0.1, 0.2, 0.3, 0.4, 0.5, 0.6, 0.7$, degree exponent $\gamma \in 2, 2.25, 2.5, 2.75, 3, 3.25, 3.5$, community size distribution exponent $\beta = 2$, average degree  $\langle k \rangle = 12$, and maximum degree $k_{max}=850$.
    \item $k$-regular networks: In a $k$-regular network, every node is connected to $k$ other nodes randomly, such that connections are random and every node has the same degree. \underline{Parameters}: $N=10000$, $k=6$.
\end{enumerate}

We note that we have left out analysis of the Watts-Strogatz (WS) model \cite{watts1998}, which is a model developed to study the small world property of random graphs. In WS, a rewiring parameter controls the trade-off between nodes clustering into triangles and the average path length between all nodes, a proxy for the small-world property. We do not study this model because (a) its degree distribution is homogeneous, following a Poisson distribution, and (b) the model does not result in networks with modular structure. Therefore, despite many uses in other contexts, the WS model is not well suited for studying the network completion problem in the present case. 

\subsubsection{Real-world Networks}
We evaluate \NOLSTAR\ on real world datasets whose characteristics are summarized in Table \ref{table:real-networks}. We show the number of nodes ($N$), number of edges ($E$), number of triangles (\#$\triangle$), and the modularity ($Q$), computed by finding the modularity maximizing partition with the Louvain algorithm \cite{louvain} with resolution parameter set to 1. Degree distribution, average clustering by degree and component size distribution is shown for each network in Figure \ref{fig:distributions}.

\begin{table}[]
\caption{Basic Characterization of Real Networks}
\label{table:real-networks}
\begin{tabular}{|l|l|l|l|l|l|}
\hline
 & Type & N & E & \#$\triangle$ & Q \\ \hline
Caida & Internet Router & 26.5k & 53.4k & 36k & 0.67 \\ \hline
Cora & Citation & 23k & 89k & 78k & 0.79 \\ \hline
DBLP & Coauthorship & 6.7k & 17k & 21k & 0.89 \\ \hline
Enron & Email Communication & 36.7k & 184k & 72k & 0.62 \\ \hline
Twitter & Social interaction & 90k & 117k & 9.4k & 0.86 \\ \hline
\end{tabular}
\end{table}

\subsubsection{Sampling Methods}
\NOLSTAR\ is agnostic to the method of sampling used to collect the initial sample graph, $\hG_0$. For the sake of continuity, all of the initial samples used in this paper were collected via \emph{node sampling with induction}. In this technique, a set of nodes is chosen uniformly at random, then a subgraph is induced on the nodes (i.e. all of the links between them are included in the sample). Our samples are defined in terms of the proportion of the edges in the underlying network. To generate samples with the target proportion of edges, we choose a sample of nodes and induce a subgraph on them; if this subgraph has too many (or too few) edges, we repeat with a larger (or smaller) subset of nodes until we find an induced graph with an acceptable number of edges.

In the main text of this paper we present results on samples collected via node sampling with induction, but we have also verified many of the results using random walk with jump sampling \cite{Ahmed2014}, which can be found in
\cref{sec:appendix} (\cref{fig:app-cumulative-reward}).

\subsection{Features}
To accurately predict the number of unobserved neighbors of a partially observed node, \NOLSTAR\ algorithms require interpretable node features that are relevant across a variety of very different network structures.  These features must be feasible to compute and update online for our algorithms to be scalable.  We use the following features for each node $i$ visible in the sample network:\\
\vspace{-0.2cm}
\begin{itemize}
\item $\hat d(i)$: the normalized degree of node $i$ \emph{in the sample network}, which can be seen as a measure of the node's centrality in the sample.  The inclusion of this feature assumes that the degree of a node in the sample is relevant to its total degree.
\item $\hat{cc}(i)$: the clustering coefficient of node $i$ \emph{in the sample network}.  This feature captures the local neighborhood density of a node, particularly the tendency of the node and its neighbors to form \emph{triangles}.
\item $CompSize(i)$: the normalized size of the component \emph{in the sample network} to which node $i$ belongs, which can be used to facilitate exploration and exploitation based on where in the network the node is located.
\item{$pn(i)$}: the fraction of node $i$'s neighbors which have already been probed.  This feature indicates the extent to which the neighborhood of node $i$ has been explored by the algorithm.
\item{$LostReward(i)$}: the number of nodes that first connected to node $i$ by being probed.  Unlike $pn(i)$, $LostReward(i)$ only counts nodes that were not connected to $i$ before they were probed. This feature mitigates the ordering effects of probing nodes. If nodes $i$ and $j$ both have the same unobserved neighbors, for instance, probing $j$ first would normally lower the total reward of node $i$ when $i$ would be probed.  $LostReward(i)$ gives credit to $i$ upon its probing, since it could have brought in as many new nodes as $j$. \footnote{\footnotesize We do not include this feature for binary reward functions, since the reward value of a query is order-independent.}
\end{itemize}

\subsection{Baseline Methods}
We compare the performance of \NOLSTAR\ with 4 heuristic baseline methods and a multi-armed bandit method.  Our heuristic baselines are as follows:
\begin{enumerate}
  \item High Degree: Query the node  with maximum observed degree to probe in every step. This has been shown to optimal in some heavy tailed networks \cite{Avrachenkov2014}.   
  \item High degree with jump: With probability $\epsilon$, query a node chosen uniformly at random from all partially observed nodes. With probability $1-\epsilon$, query the node with the maximum observed degree. We set $\epsilon=0.3$.
  \item Random Degree: Query a node chosen uniformly at random from all partially observed nodes.
  \item Low Degree: Query the node with minimum observed degree in every step. This method is approximately\footnote{\footnotesize Cases where low degree is not optimal, even in networks with uniform degree, can be constructed. Specifically, if the neighbors of the node with lowest degree are already in the network, querying it will result in reward 0. If in the same case the neighbors of the node with 2nd lowest degree are almost all outside of the network, reward for querying that node will be larger.} optimal for $k$-regular networks where every node has the same degree, since the lowest degree node is always furthest from the uniform degree $k$ (see \cref{sec:appendix}(\cref{fig:app-er-reg})). 
\end{enumerate}

We also compare our approach with the nonparametric multi-armed bandit model proposed in \cite{madhawa2019multiarmed}, which we refer to as KNN-UCB (k-nearest-neighbors upper confidence bound). This method similarly relies on computing a vector of structural features for each node, including degree, average neighbor degree, median neighbor degree and average fraction of probed neighbors. Each unprobed node is considered an arm in a Multi-armed Bandit formulation, and the next arm to pull (node to probe) is chosen by computing $argmax_i{\hat{f}(x_i) + \alpha \sigma{(x_i)}}$.
 Here, $\hat{f}(x_i)$ is the expected reward of probing node $i$ and $\sigma (x_i)$ is the average distance to other points in the neighborhood. The expected reward is calculated as a weighted k-NN regression on nodes within the \textit{k-NN radius} of node $i$, defined as the $k$ nodes whose feature vectors have Euclidean distance less than $r$ from the feature vector of node $i$. The term $\alpha \sigma(x_i)$ facilitates exploration by allowing the possibility of nodes without maximum expected reward to be probed, assuming $\alpha > 0$. In our experiments we fixed the value of $k=20$ and $\alpha=2$, following the experiments in the original paper.
 

\subsection{Experimental Setup}
Across all networks, our experiments are run over 20 independent initial node samples of the network. In synthetic networks, the underlying network is a realization of the model using the parameters described in Section \ref{sec:synthetic}. 

Our experiments aim to (1) investigate how network properties impact the performance of \HTR, (2) exhibit the performance tradeoffs between \HTR\ and \NOLOR, (3) show that \NOLSTAR\ algorithms outperform the baseline methods in settings where learning is possible and approximate the heuristic methods elsewhere, (4) analyze the prediction error of \HTR, and (5) analyze the evolution of the feature weights learned by \HTR\ over time.

\begin{figure*}[!ht]
\centering
\begin{subfigure}{0.33\textwidth}
\centering
	\includegraphics[width=\linewidth, height=3.5cm]{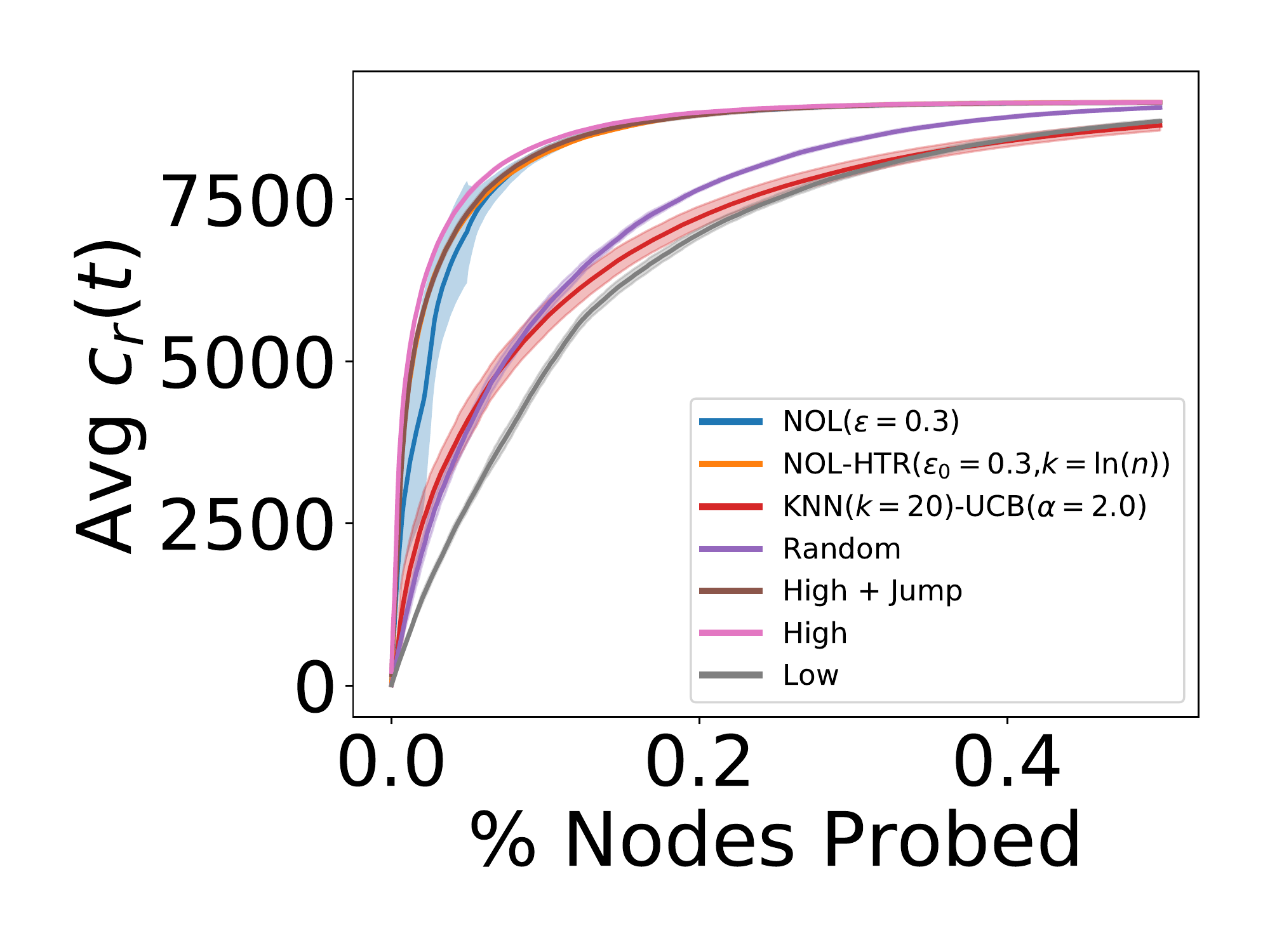}
    \vspace{-0.7cm}
    \caption{BA}
    \label{fig:cumulative-reward-BA}
\end{subfigure}%
\begin{subfigure}{0.33\textwidth}
	\includegraphics[width=\linewidth, height=3.5cm]{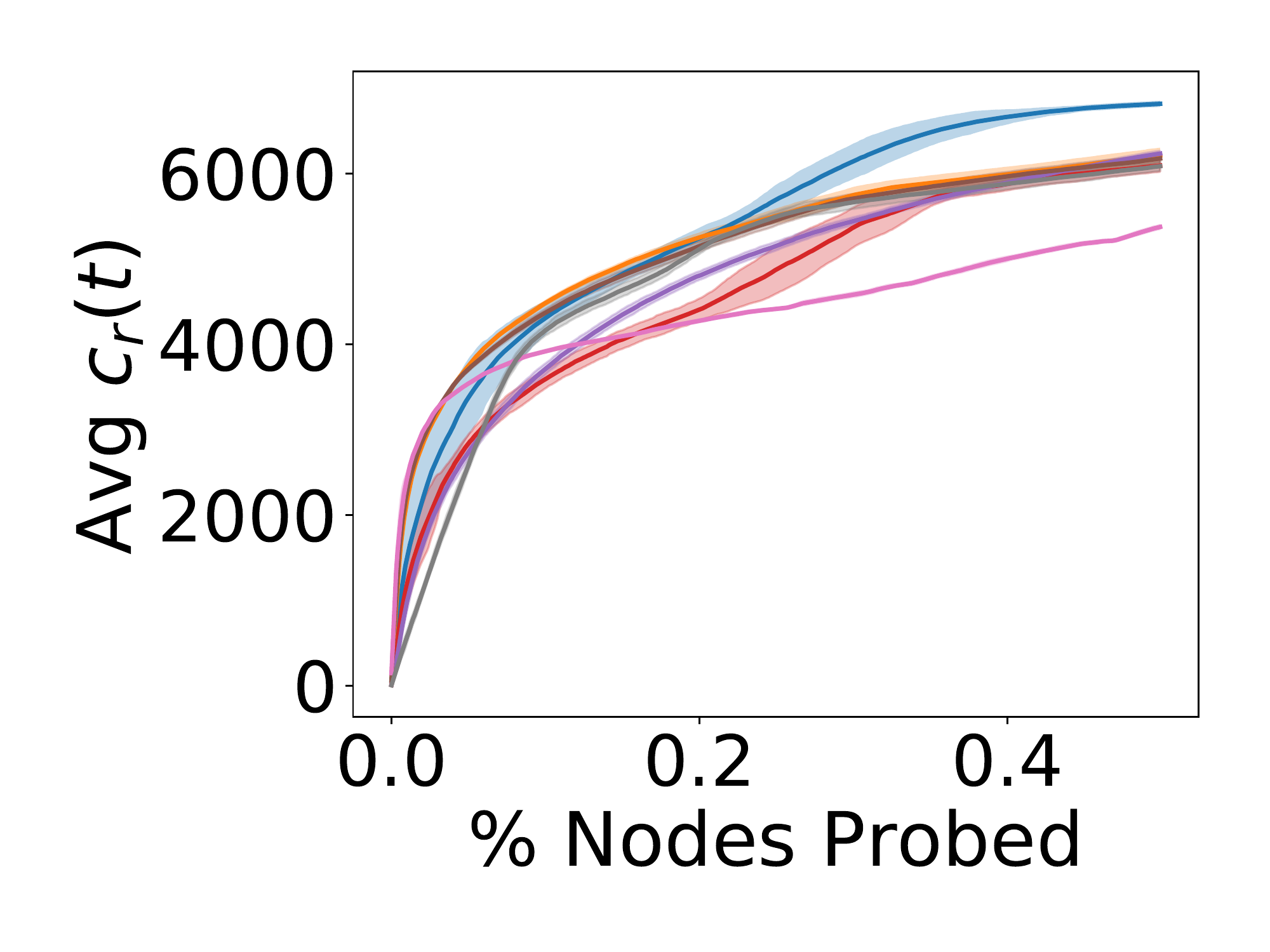}
    \vspace{-0.7cm}
    \caption{BTER}
    \label{fig:cumulative-reward-BTER}
\end{subfigure}%
\begin{subfigure}{0.33\textwidth}
	\includegraphics[width=\linewidth, height=3.5cm]{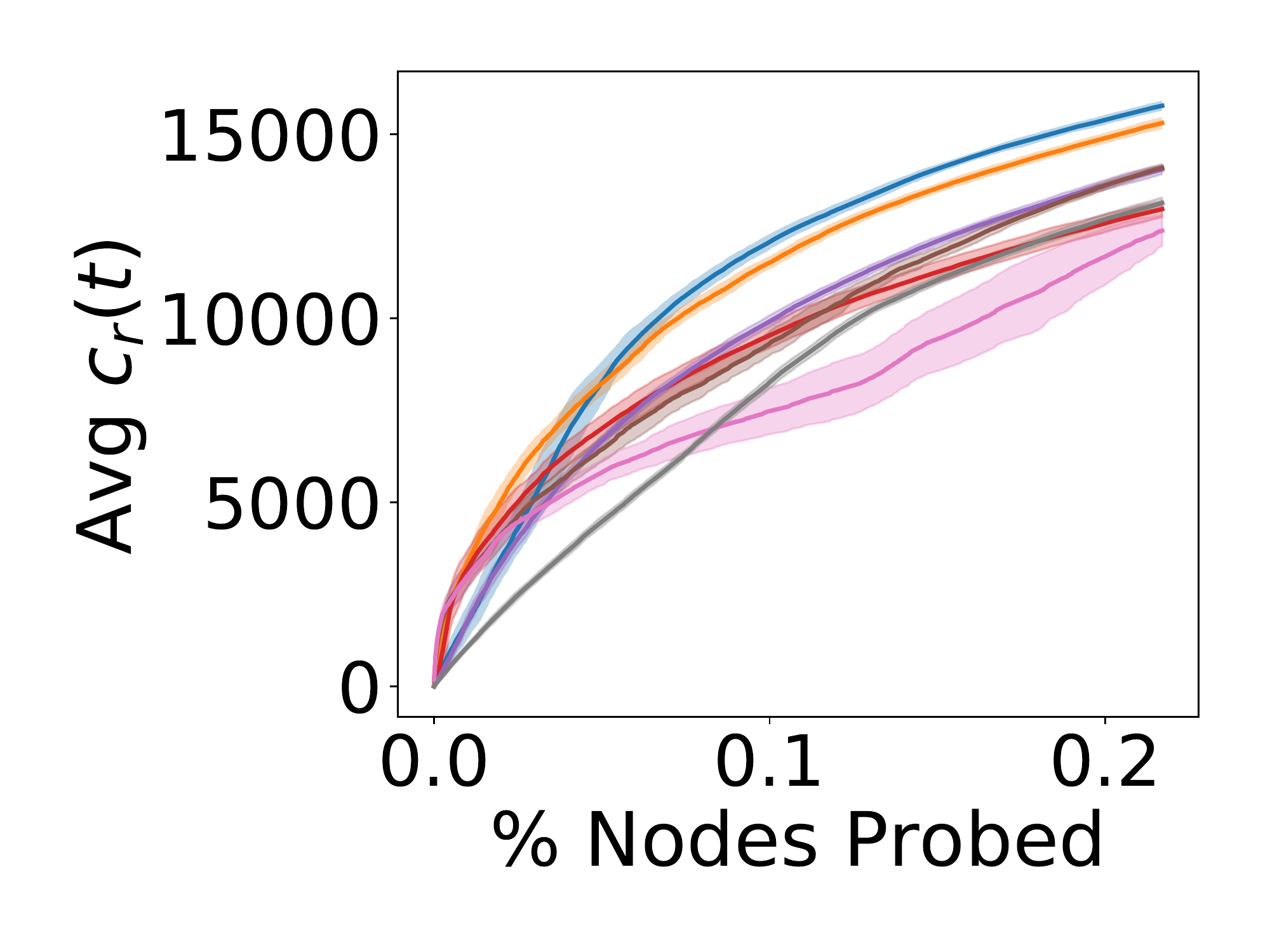}
    \vspace{-0.7cm}
    \caption{Cora}
    \label{fig:cumulative-reward-cora}
\end{subfigure}
\begin{subfigure}{0.33\textwidth}
	\includegraphics[width=\linewidth, height=3.5cm]{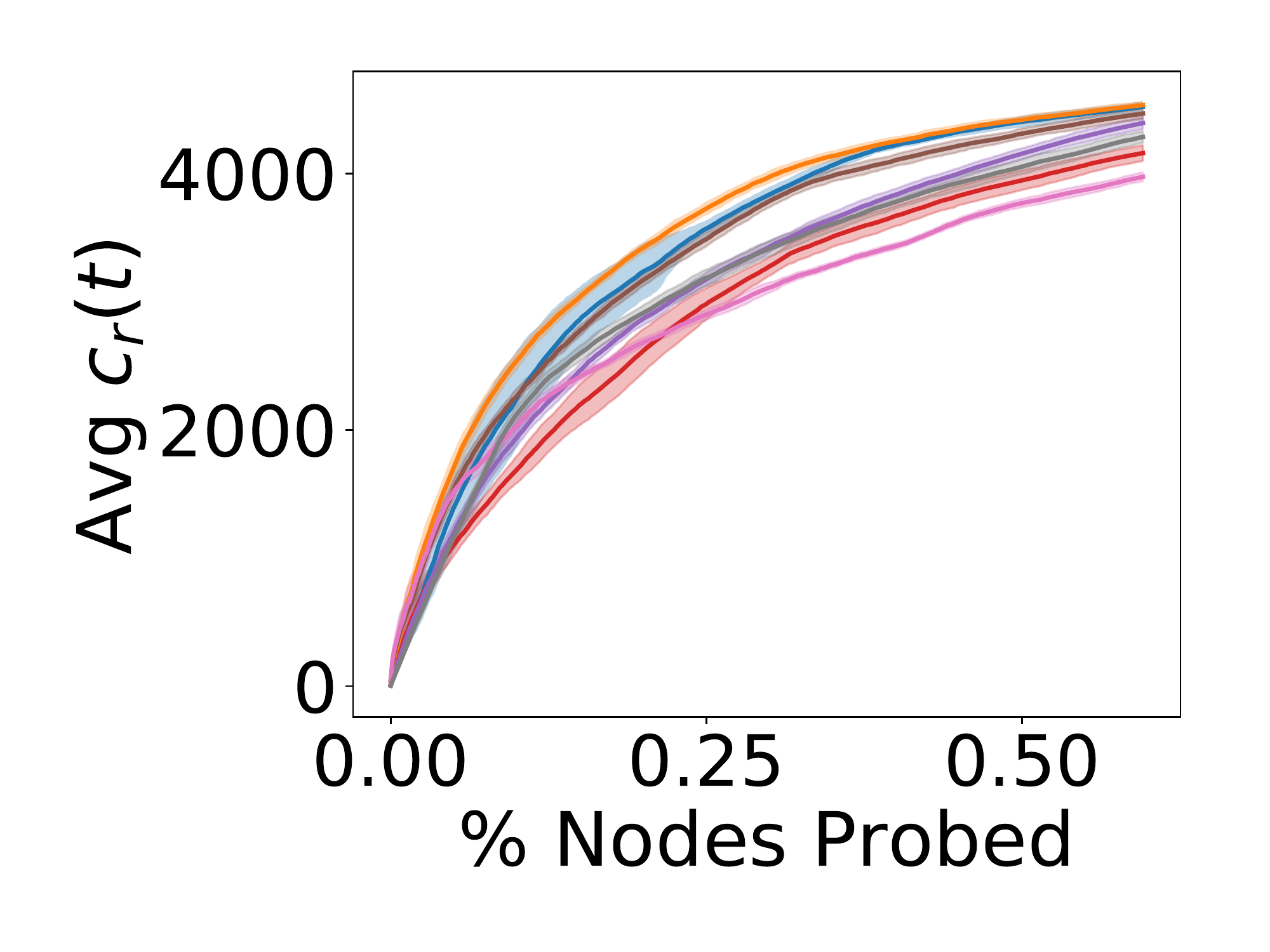}
    \vspace{-0.7cm}
    \caption{DBLP}
    \label{fig:cumulative-reward-dblp}
\end{subfigure}%
\begin{subfigure}{0.33\textwidth}
	\includegraphics[width=\linewidth, height=3.5cm]{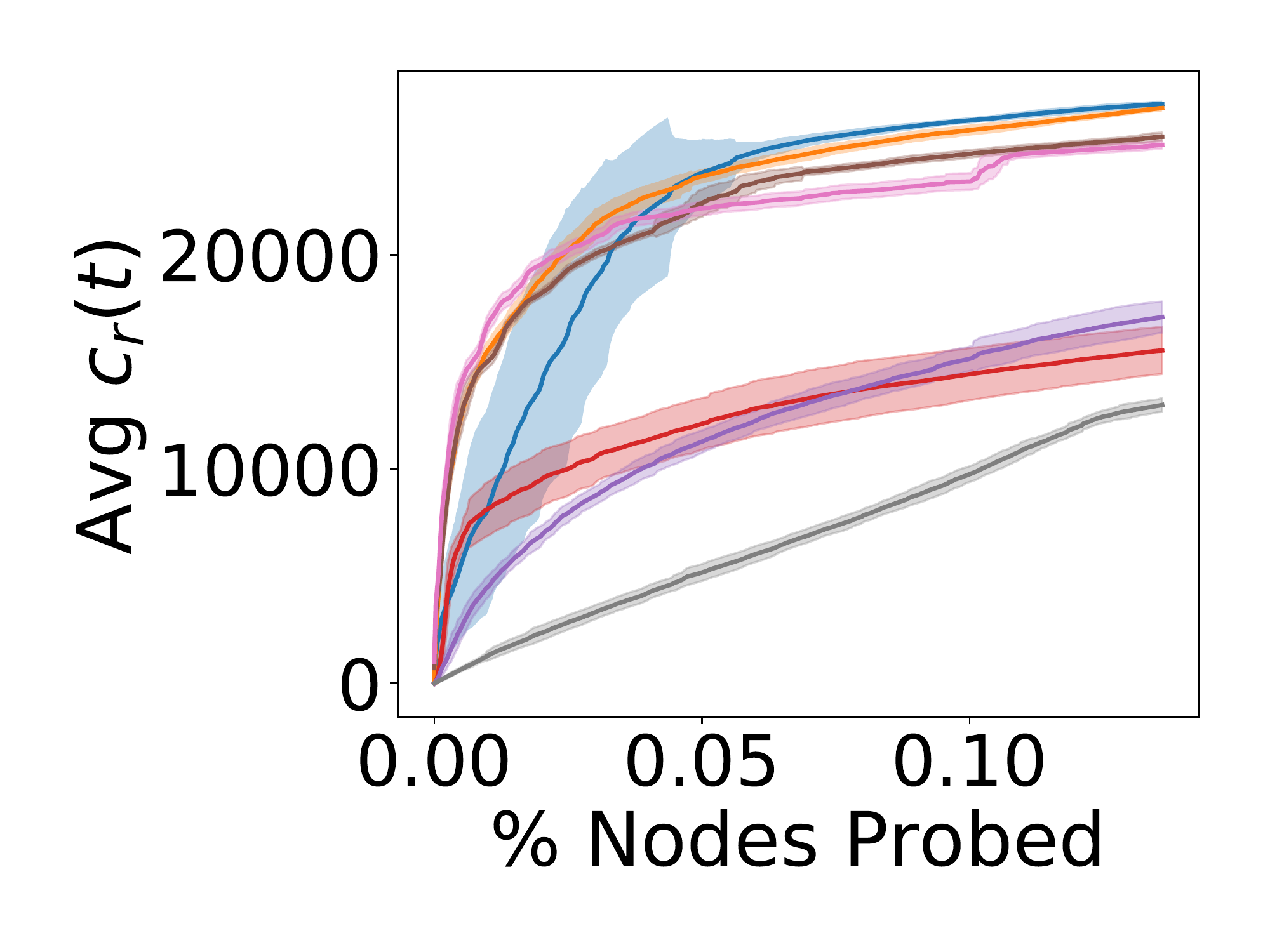}
    \vspace{-0.7cm}
    \caption{Enron}
    \label{fig:cumulative-reward-enron}
\end{subfigure}%
\begin{subfigure}{0.33\textwidth}
	\includegraphics[width=\linewidth,height=3.5cm]{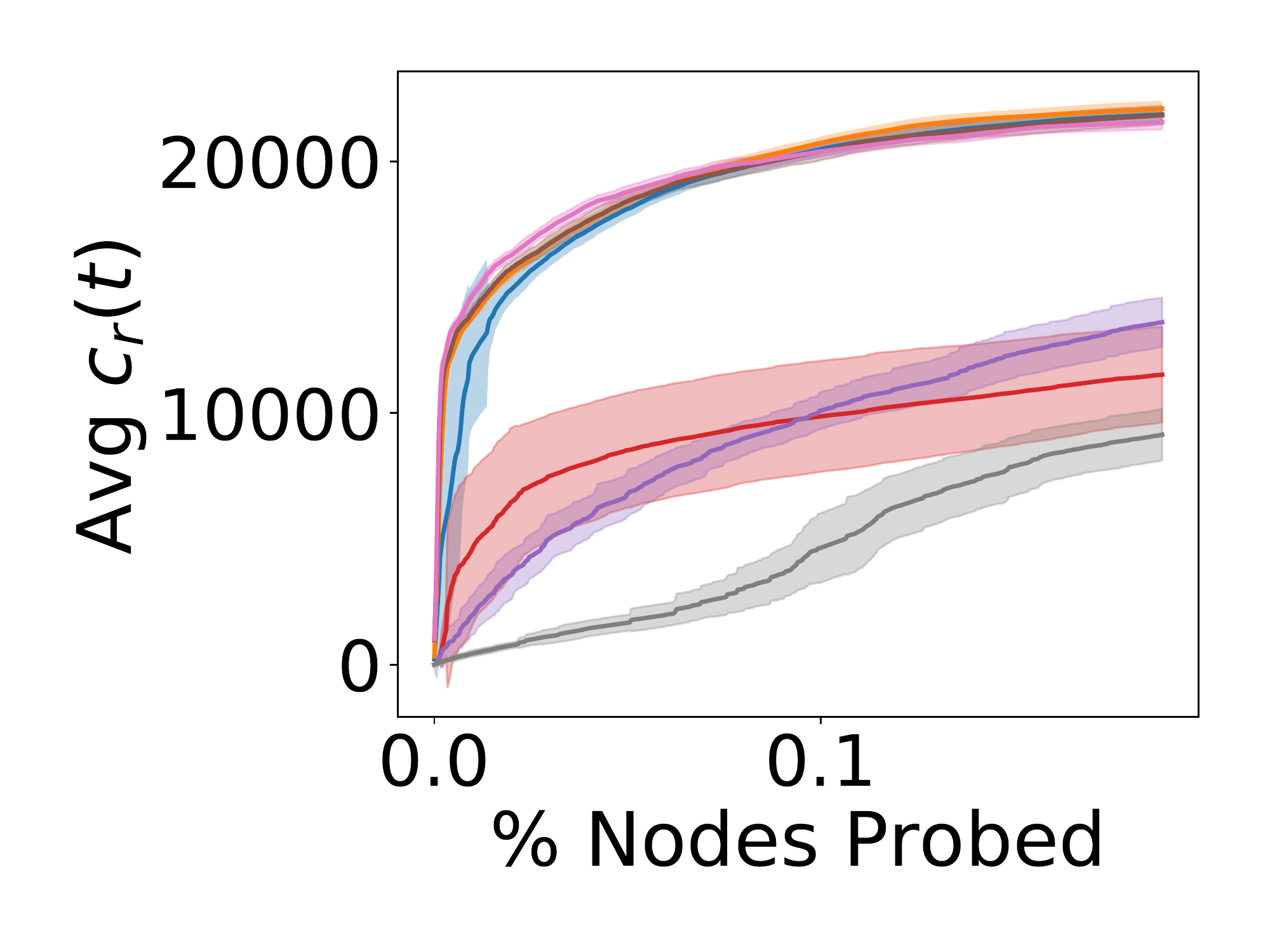}
    \vspace{-0.7cm}
    \caption{Caida}
    \label{fig:cumulative-reward-caida}
\end{subfigure}
\vspace{-0.3cm}
    \captionsetup{width=0.9\textwidth}
    \caption{Cumulative reward, $c_r(t)$, averaged over experiments on multiple independent samples of both synthetic (BA, BTER) and real world networks. In the BA network (a), where probing the highest degree node is optimal, NOL "learns the heuristic". NOL outperforms the baseline methods in BTER networks (b), where the combination of heavy tailed degree distribution and relatively high clustering and modularity allows for discrimination. In some real networks (c, d, e), NOL either outperforms or closely tracks the best baseline, while in a real network with properties similar to a BA model (f) NOL is outperformed by the high degree baseline.}
    \label{fig:cumulative-reward}
\end{figure*}

\subsection{Performance Metrics}
\label{subsec:performance}
After each probe of the network, \NOLSTAR\ earns a \emph{reward}, defined in this work as the number of previously unobserved nodes included in the network after a probe. Formally, the reward at time $t$ is defined as $r_t = \left|V_{t+1}\right| - \left|V_{t}\right|$. 
  We study the performance of each method by showing the \emph{cumulative reward}, $\hat{c}_r(T)$, where $T$ is a time step between $0, 1, 2, \dots b$. Formally, $\hat{c}_r(T) = \sum_{t = 1}^T r_t$.
 We also want to quantify the utility of decisions made by \NOLSTAR. For this purpose, we study the prediction error of the model. This quantifies the extent to which our prediction, $\V_{\theta_t}(\phi_t(i))$, differs from the true reward value $r_t$. Thus, we calculate $E(t) = \V_{\theta_t}(\phi_t(i)) - r_t $.

\subsection{\HTR\ Parameter Search}
\label{sec:param-search}
We ran a two dimensional parameter search over values of $k$ (1-16, 32, 64, 128, $\log_{10}(n)$, $\log_e(n)$, $\log_2(n)$) and $\epsilon$ (0, 0.1, 0.2, 0.3, 0.4, and exponential decay versions of each). We ran this search on all of the networks in \cref{fig:distributions}), as well as 56 LFR networks \cite{LFR} spanning a wide variety of network structures. We varied two parameters in the LFR networks: $\mu$, which controls modular structure by adjusting the probability of cross-community links; and $gamma$, which controls the exponent of the degree distribution of the entire network \cite{LFR}. 

Across all of these networks, we did not observe a general trend in which some parameter settings performed best consistently across experiments. There was not a single best choice or regime of parameters through the experiments for any of the networks we searched on individually, nor was there a standout choice of parameter across the networks.

However, we have found some evidence to suggest that performance on more modular structures is improved by more randomization, meaning a non-zero value of $\epsilon$. We ran a linear regression using $\epsilon$ as the target variable and global clustering, modularity, and degree exponent as covariates. All three covariates positively effected $\epsilon$, with global clustering having the strongest effect (2.33). This result is consistent with a network scientific understanding of the querying process: when clustering is relatively prevalent, the likelihood of finding a local minima within one well clustered community is high, meaning that adding more randomness to the algorithm will increase the likelihood that the algorithm sees examples that allow it to avoid such a minima.

We report results using a set of parameters that performed reasonably well across networks. These parameters are $\epsilon=0.3$, decaying as $\epsilon_t = 0.3 e^\frac{-t}{b}$ and $k=\ln(t)$, where $t$ corresponds to the number of nodes probed thus far in the experiment.

\subsection{Results}
In this section, we compare the performance of the heuristic baseline methods, the KNN-UCB baseline approach, \NOLOR, and \HTR. Broadly, we find that \HTR\ and \NOLOR\ perform similarly in terms of average cumulative reward, but that the performance of \HTR\ is more consistent, indicated by standard deviations that are tighter around the mean across experiments.

\paragraph{Cumulative Reward}
Figure \ref{fig:cumulative-reward} shows average cumulative reward $\hat{c}_r(T)$ over a budget of thousands of probes on 6 networks. The average and standard deviation of $\hat{c}_r(T)$ are computed over experiments on 10 independent samples. We omit results on ER networks, noting that because all nodes are statistically equivalent in terms of structural properties, every probing method performs equivalently and neither learning or heuristics provide any significant advantage (see \cref{fig:app-er-reg}).

 In BA networks (Figure~\ref{fig:cumulative-reward-BA}), \HTR\ performs on par with the High Degree baseline, which is known to be near optimal in networks with heavy tailed degree distributions \cite{Avrachenkov2014}. Further, \HTR\ outperforms \NOLOR by achieving both higher average reward and smaller standard deviation. 
 \HTR\ also consistently outperforms the baseline methods in networks generated by the BTER model (Figure \ref{fig:cumulative-reward-BTER}). Although \HTR\ outperforms \NOLOR\ in both reward and variance towards the beginning of the experiment, after around 25\% of the nodes have been probed \NOLOR\ begins to outperform \HTR. See Section \ref{subsec:tradeoffs} for a discussion of some potential explanations for this observation. 
 \HTR\ performed as well as \NOLOR\ and the best heuristics in every real world network we experimented on. Comparing directly with \NOLOR, \HTR\ is able to achieve similar or better performance, always with smaller standard deviation, on every network, regardless of the underlying distributions (compare networks in Figure \ref{fig:distributions}). 
 We note that across experiments, the KNN-UCB method was unable to match the performance of our model and often underperformed the baseline methods.\footnote{\footnotesize We further note that the sample collection techniques used in the experiments in the KNN-UCB preprint \cite{nonparametric} varied significantly from those we employ here and that all of our experiments ran over larger probing budgets.}

\begin{figure}
	\includegraphics[scale=0.4]{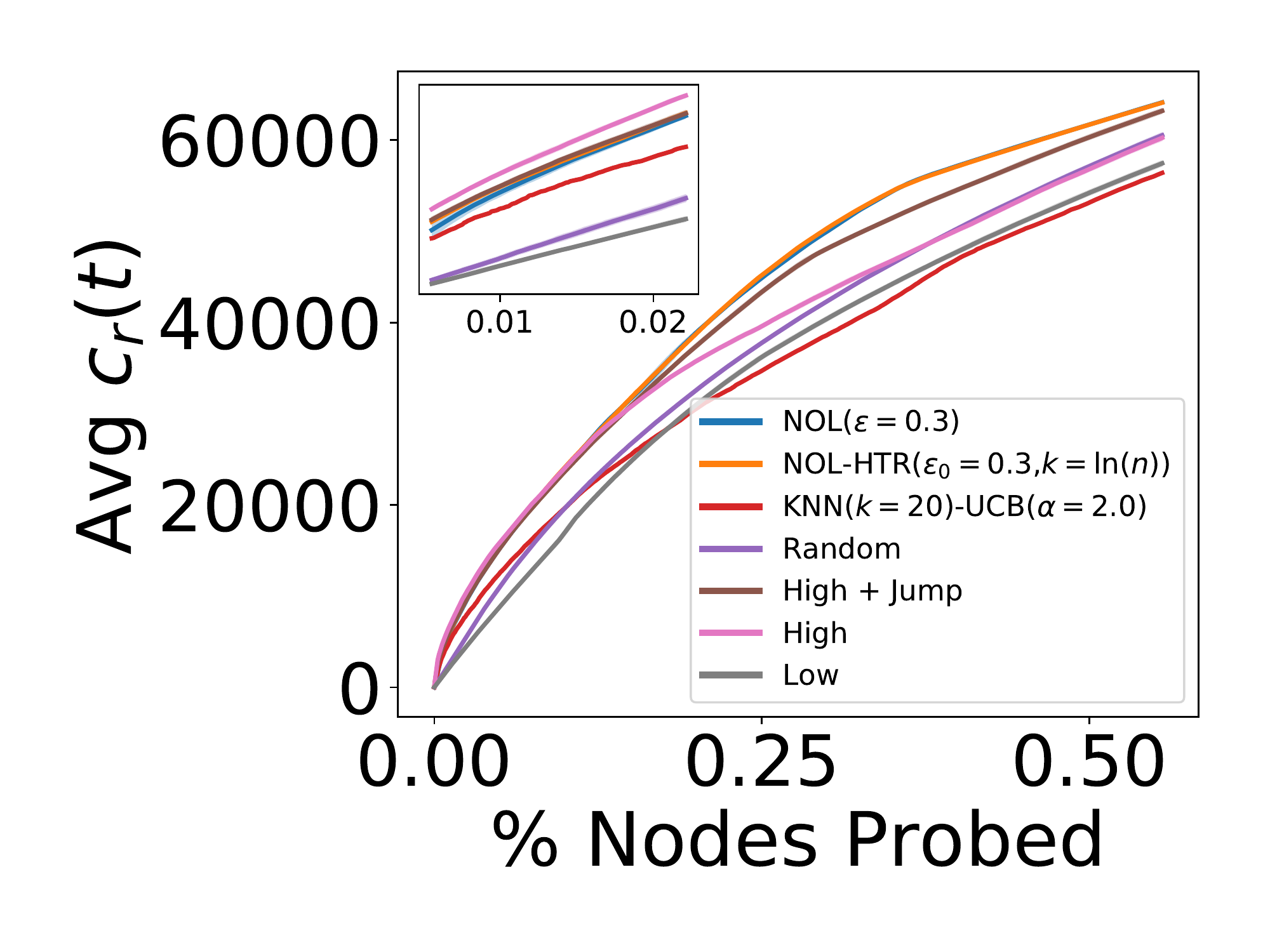}
	\captionsetup{width=0.9\textwidth}
	\caption{Cumulative reward results on a Twitter social network constructed from mutual interactions. The budget in the experiment was 50k queries. Twitter data is typically collected via repeated API access, so represents a natural use case for \HTR. Inset: Magnification showing that \HTR\ and the high degree baseline methods outperform \NOLOR\ in the first few thousand probes of the experiment.}
	\label{fig:twitter}
\end{figure}

\paragraph{Use Case: Twitter Social Network}
Figure \ref{fig:twitter} shows results on a social interaction network sampled from Twitter. In this setting, querying a node corresponds to calling the Twitter API to obtain more data about a specific account. This is an example of a natural use case for \HTR, both because the network must be expanded by repeatedly accessing an API and because there is evidence that the distribution of social interactions is heavy-tailed (see e.g. \cite{watts2006,Benito2012}). 

This Twitter Social Network data set consists of Twitter users connected to one another with an undirected edge if they mutually retweeted or mentioned each other throughout the course of a single day in 2009. We focus our experiment on the largest connected component of this interaction network. While this is only a small fraction of the Twitter network, the data was collected via the Twitter Firehose, and is therefore complete for that time span, which allows us to accurately simulate the process of growing a network by querying Twitter users. Some other characteristics of the dataset are outlined in Table \ref{table:real-networks} and the distributions of degree and clustering are shown in Figure \ref{fig:distributions}. Notably in our context, though the maximum degree in this network is not very large ($\approx 100$), choosing a random node with degree near the maximum is much less likely than choosing a node with low degree. 

As shown in Figure \ref{fig:twitter}, the \NOLOR\ method outperforms the heuristic baseline methods and \HTR\footnote{\footnotesize Our attempt to run the KNN-UCB baseline on the Twitter network did not finish in a reasonable amount of time, therefore we omit it.}. However, the inset plot shows that, similar to the experiments on other networks, \HTR\ outperforms \NOLOR\ early in the experiment before eventually being overtaken. Next, we discuss our conjectures about why we observe this trade-off behavior.


\paragraph{Prediction Error}
We analyze the ability of \NOLSTAR\ to learn by showing the measure of \emph{prediction error}, $E(t)$, defined in Section \ref{subsec:performance}, over time and across different initial sample sizes.

\begin{figure}[ht]
	\centering
	\includegraphics[width=0.45\linewidth]{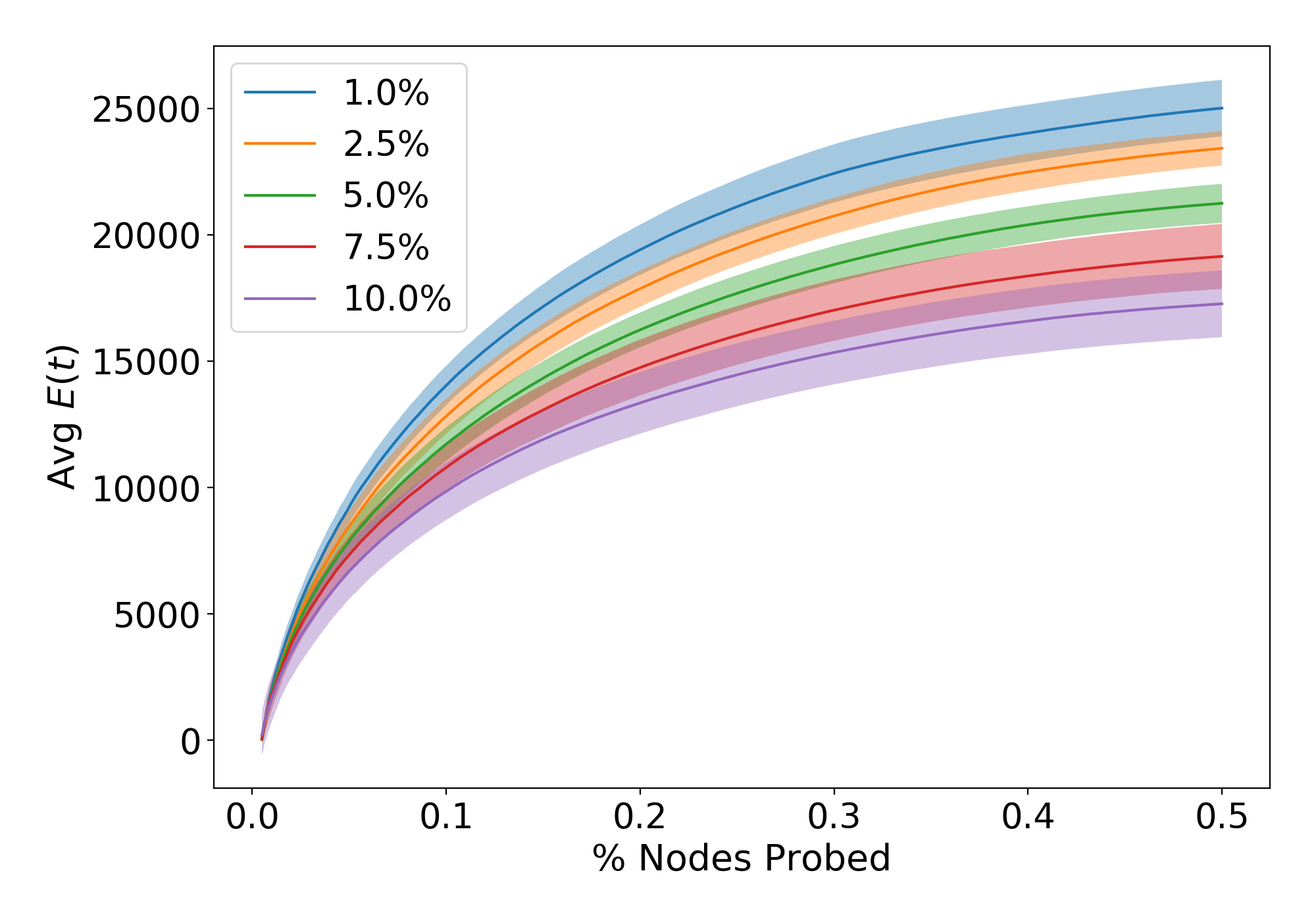}%
	\includegraphics[width=0.45\linewidth]{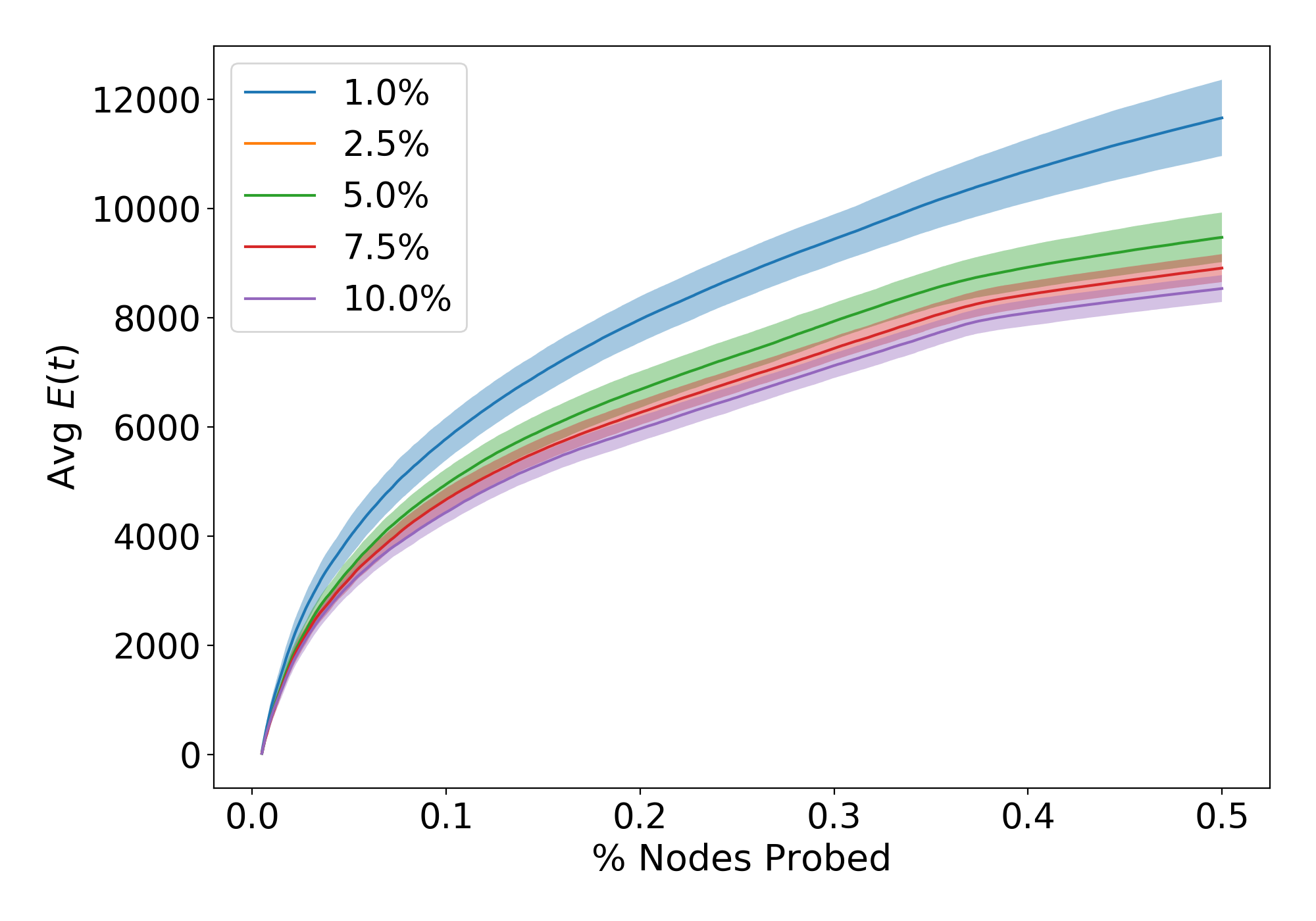}
	\captionsetup{width=0.9\textwidth}
	\caption{Average cumulative prediction error of \HTR\, starting at time step $t=50$. Smaller initial samples result in larger prediction error, but error stabilizes over probes regardless of initial sample size.}
	\label{fig:avg_PE_plots}
\end{figure}

Figure \ref{fig:avg_PE_plots} shows the cumulative prediction error, $E(T)$, as a function of time. The error is averaged over 10 independent samples of the network for initial sample sizes of 1\%, 2.5\%, 5\%, 7.5\% and 10\% of the complete network (as percent of edges in the network). 

We show results in both the BA and BTER models. The predictions of \HTR\ are noisy in the beginning of the experiment and the resulting outliers skew the analysis of cumulative error, therefore we present the error starting from $t=50$. In both cases the prediction error is relatively stable over time, indicated by the slow growth of the curves. We also observe that the average prediction error is lowest when the initial sample is largest, and highest when the initial sample is smallest; This is intuitive, since when the initial sample is large, the model is learning its predictions from more accurate information, whereas when the initial sample is small, the training information is very noisy.

%


\paragraph{Feature Weight Analysis} 
We qualitatively analyze the feature weights learned by \HTR\ over time (see \cref{fig:app-feature-weights} for visualization). Since the weights were computed on random subsamplings of the observed data at every time step, they fluctuated considerably between probes. Still, we found that across most networks (all but Caida and Enron), the degree of a node was weighted positively and large in magnitude, implying that it is the most salient feature to predict reward. This is intuitive, since we expect the target variable to correlate highly with the sample degree. There were no features with consistently highly negative weights, which would imply a negative predictor of reward. Instead, the other features typically fluctuated around 0, meaning they were not consistently predictive of either high or low rewards, but were non-zero so did contribute to the prediction. 

The Caida and Enron networks were exceptions to the above general trends, though \HTR\ exhibits similar performance on these networks in terms of cumulative reward. In these experiments, the weight of the degree feature was centered around 0 along with the other features, but with very large fluctuations in both positive and negative directions. This suggests that the importance assigned to degree depended strongly on the particular subsampling of the data in an individual time step. It may also have implications for the impact of degree correlations (i.e. average neighbor degree), since the fluctuations are consistent with both low and high degree nodes predicting high reward.

\subsection{Comparing NOL and NOL-HTR}
\label{subsec:tradeoffs}
The above analysis illustrates some differences and tradeoffs between \NOLOR\ and \HTR: 
\textbf{(1)} \HTR\ achieves similar performance to \NOLOR\ and is more consistent. This is evidenced by the tighter standard deviations around the average cumulative reward. 
\textbf{(2)} \HTR\ outperforms \NOLOR\ in the beginning of every experiment. This is consistent with our expectation based on how each of the algorithms work: \HTR\ is able to leverage outlier, high reward queries early on because it computes maximum likelihood parameters at every step, while \NOLOR\ updates its parameters with online gradient descent using a fixed learning rate, and is therefore not able to adapt as effectively to high reward nodes. 
\textbf{(3)} As the number of probes increases, \NOLOR\ often begins to outperform \HTR. There are a few possibilities for why this is happening. First, \NOLOR\ is learning through gradient descent, so adjusting the learning rate of the algorithm could impact the amount of examples in the tail it takes to reach highly predictive parameters, explaining the lag. Second, as the number of samples grows, the \HTR\ parameter setting of $k=ln(n)$ grows very slowly. This means that more data is being subsampled into the same number of bins, thus each subsample may become more noisy and the outliers become less distinguishable. This could be alleviated by increasing the value of $k$ more quickly as the number of samples grow, or by capping the size of the subsamples, or by choosing the sample size so that data from the tail is more likely to have an impact on the parameters.

\subsection{Limitations of Learning}
\label{subsec:limits}
We are interested in understanding the limitations of our approach in this setting. We experiment across a wide variety of network structures using the LFR community benchmark \cite{LFR}. We generated 56 LFR networks spanning two parameters: $\mu$, which controls modular structure by adjusting the probability of cross-community links; and $\gamma$, which controls the exponent of the degree distribution of the entire network. Then, we ran the \HTR\ parameter search (see \cref{sec:param-search} on each network, as well as random and high degree baseline methods.
For each network, we found the set of \HTR\ parameters that resulted in the maximum average cumulative reward and computed the percent gain in performance using either of the baselines as $$c_r^\Delta = \frac{c^{\text{HTR}}_r - c^{\text{base}}_r}{c^{\text{base}}_r} \times 100$$

\begin{figure}
    \centering
    \captionsetup{width=0.9\textwidth}
    \includegraphics[width=0.95\textwidth]{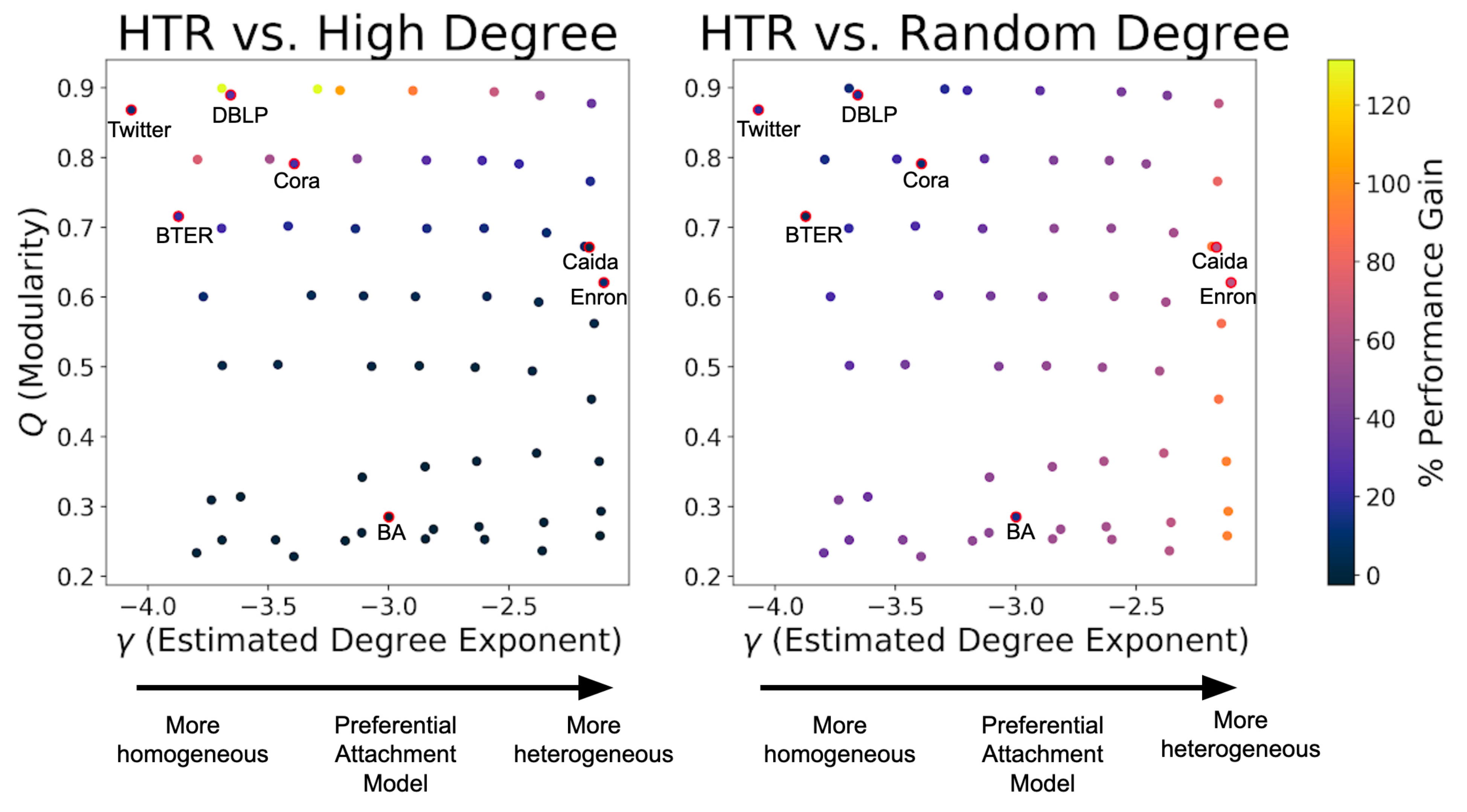}
    \caption{Performance gain using best performing \HTR\ parameters compared to High Degree (left) and Random Degree (right) heuristics on real network data and LFR benchmark graphs \cite{LFR} with varying degree heterogeneity and modular structure. \HTR\ almost always outperforms both heuristics, but the gains are more substantial depending on the network structure. Performance over High Degree improves most when the network is modular with a more homogeneous degree distribution, while performance over Random Degree improves most when the degree distribution is very heterogeneous. }
    \label{fig:lfr-scatter}
\end{figure}

Results are shown in \cref{fig:lfr-scatter}. Compared with high degree probing (left plot), \HTR\ is always able to gain in performance in higher modularity networks ($Q>0.6$), with performance gains spanning from 15-30\% up to about 130\%. When modularity is lower, including in the BA model realization, \HTR\ approximates high degree across degree exponents, with the maximum performance loss of less than 3\% (-2.54\%) compared to the heuristic. 

Comparing with random degree probing (right plot), \HTR\ is always able to gain in performance, with minimum performance gain of 5\% and maximum of 93\%. The maximum gains are in networks with the most heterogeneous degree distributions. This corresponds to the most star-like network structures, meaning most randomly chosen nodes will have very low degree. 

The performance compared to random degree probing in the real networks seem to fit with performance on LFR networks with similar parameters, allowing for some noise in either direction.

Comparing with High Degree on BTER, Caida, Cora and Enron, performance appears to be about on par with similar LFR networks, again allowing for some noise.

Performance gain by using \HTR\ rather than High Degree in the DBLP network is substantial, but smaller than an LFR network with similar properties.

The degree exponent estimate for our Twitter sample is an outlier in this dataset. However, there is also substantial performance gain, though it might be smaller than what we would expect given an LFR network with similar properties.
\section{Conclusion and Future Work}
\label{sec:concl}
We proposed and evaluated algorithms to address the problem of reducing the incompleteness of a partially observed network via successive queries as an online learning problem.  We presented two algorithms in the \NOLSTAR\ family and highlighted \HTR\ for learning heavy-tailed reward distributions. We showed that \HTR\ is able to consistently outperform other methods, especially early-on in the process of querying nodes when the extreme values are yet to be discovered. We also showed that macroscopic properties of the underlying network structure, specifically the degree distribution and extent of modularity, are important factors in understanding when learning will be relatively easy, difficult, or nearly impossible. Alongside experiments on multiple synthetic and real world networks, we presented experiments on a Twitter interaction network, a realistic use case for a network growth algorithm such as \HTR. 

The problem of online network discovery remains fruitful for future work. The case of noisy observations from the query model has yet to be fully addressed. For example, a query may return only a sample of the neighbors, or a list of potential neighbors that may include false positives. The specific models we presented here are not sensitive to this type of noise, but could be addressed in extensions to \NOLSTAR. Similarly, an adversarial version of the problem can be formulated, where an adversary is intentionally poisoning the queries (via either the query that is sent or the data that is returned) and the model must include the adversary in order to make appropriate adjustments to decision making. Such noisy network discovery tasks could be formulated as Partially Observed Markov Decision Processes (POMDPs), which differ from our MDP formulation by the fact that the agent is uncertain about its current state (for example, we could model some error on the feature vectors). 


\begin{backmatter}

\section*{Availability of data and material}
Code and data to reproduce the results in this paper are available at \url{https://github.com/tlarock/nol/}. Datasets used in our experiments were downloaded from SNAP \cite{snapnets}.

\section*{Competing interests}
The authors declare that they have no competing interests.

\section*{Funding}
This research was sponsored in part by (1) the National Science Foundation grants NSF CNS-1314603 and NSF IIS-1741197, (2) the Combat Capabilities Development Command Army Research Laboratory and was accomplished under Cooperative Agreement Number W911NF-13-2-0045 (ARL Cyber Security CRA), and (3) the Under Secretary of Defense for Research and Engineering
under Air Force Contract No. FA8702-15-D-0001. The views and conclusions contained in this document are those of the authors and should not be interpreted as representing the official policies, either expressed or implied, of the Combat Capabilities Development Command Army Research Laboratory or the Under Secretary of Defense for Research and Engineering or the U.S.~Government. The U.S.~Government is authorized to reproduce and distribute reprints for Government purposes not withstanding any copyright notation here on.

\section*{Author's contributions}
This manuscript would not exist without TL; he was the lead contributor in all aspects of the manuscript. TS was involved in generating the synthetic data sets, preparing the Twitter data, and generating the feature importance results.  SB was involved in the design of algorithms presented in the manuscript.  TER supervised the work and contributed to methodology and experimental design.  TL was the lead developer of the code and conducted all the experiments.  TL was the lead contributor in writing the manuscript. All authors read and approved the final manuscript.


\bibliographystyle{bmc-mathphys}
\bibliography{references}

\clearpage
\appendix
\section{Appendix}
\label{sec:appendix}

\subsection{Results on ER and Regular networks}
In \cref{fig:app-er-reg}, we present cumulative reward results on an ER network (left) and a k-regular network (right). 
The result on the ER network shows that every querying strategy performs indistinguishably from the others. 
This is due to the fact that the degree distribution of the network is homogeneous, meaning that the expected number of neighbors of each node is well defined and the same for all nodes. 
Since the expected number of neighbors is the same for every node, but the exact number of neighbors for a given node is random, no querying strategy is able to outperform any other. 

In the k-regular network, every node has exactly the same degree, but the particular neighbors are randomly chosen.
Since every node has the same degree, choosing the lowest degree node is approximately optimal when maximizing the number of newly observed connections.
For the same reason, choosing the highest degree node is usually far from optimal, since the maximum degree node in the sample will have degree nearly k, thus the minimum reward.
This explains why high degree is the worst performer.


\begin{figure}[ht]
\centering
\includegraphics[width=0.5\linewidth]{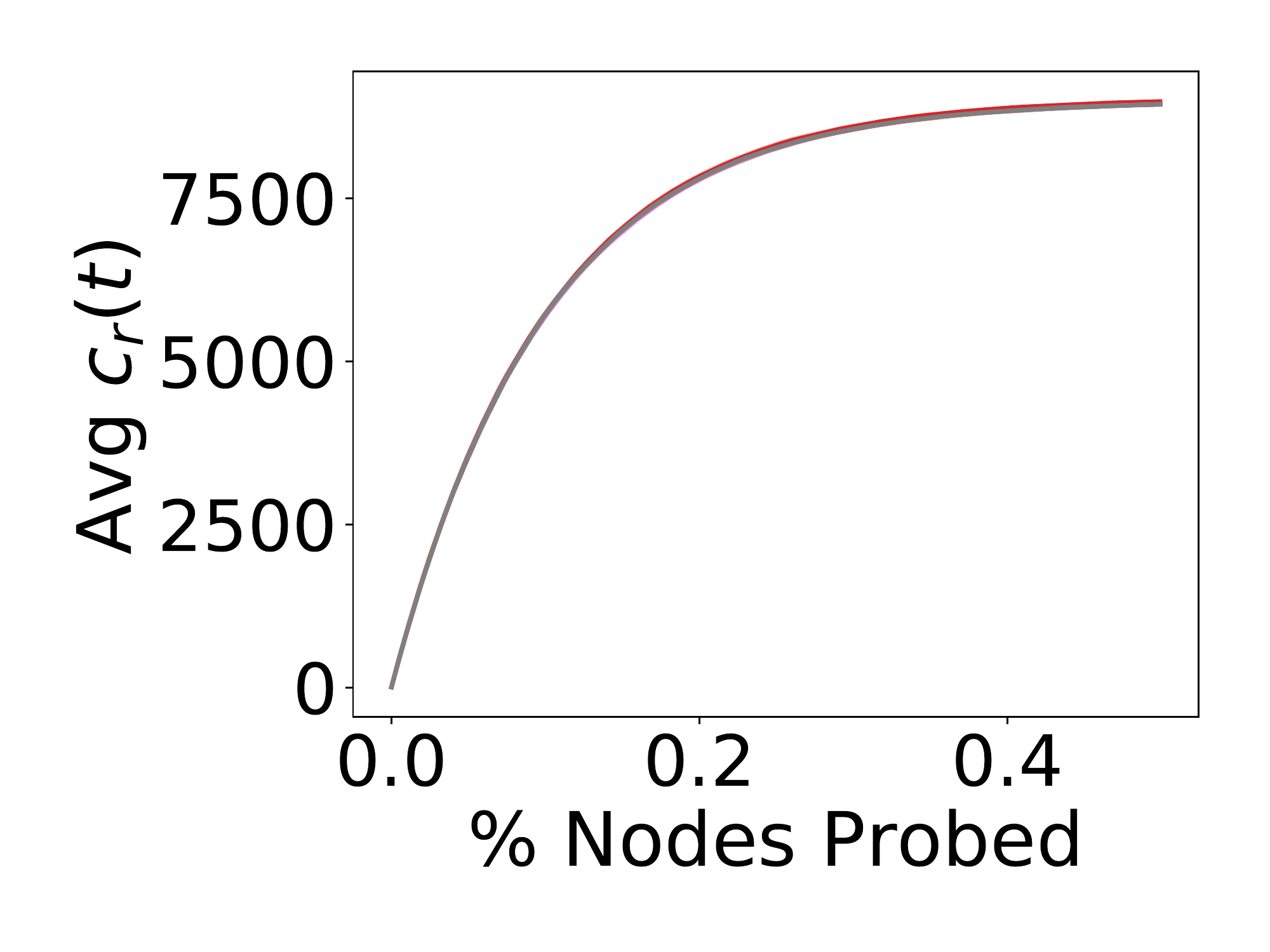}%
\includegraphics[width=0.5\linewidth]{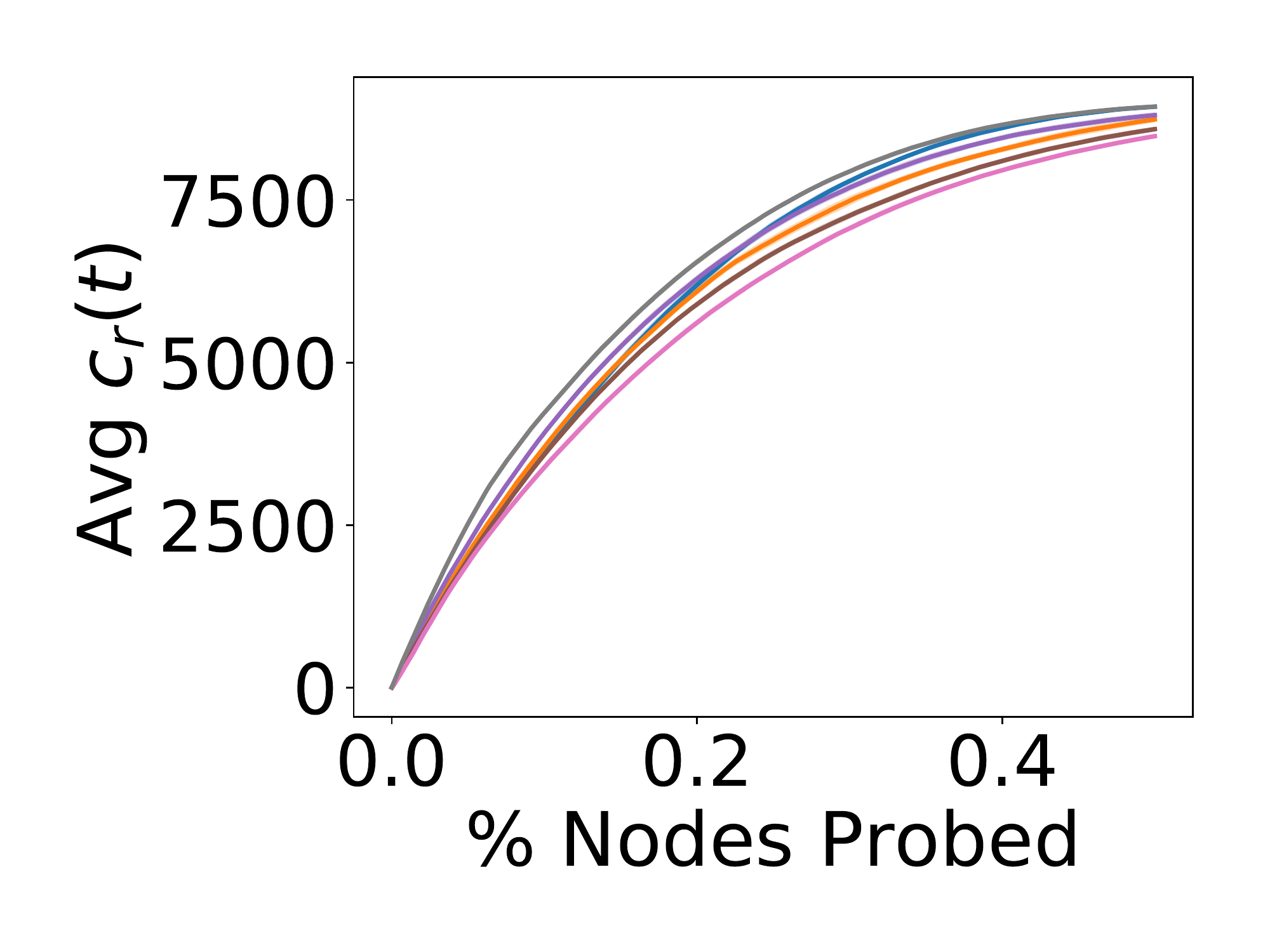}
\vspace{-0.8cm}
\captionsetup{width=0.9\textwidth}
\caption{Results on ER and regular networks. All methods perform indistinguishably in ER networks because the degree distribution is homogeneous and there is little clustering or modular structure. Querying the lowest degree node is approximately optimal in the regular network, since the lowest degree node has the maximum missing connections.}
\label{fig:app-er-reg}
\end{figure}

\subsection{Feature Weight Analysis}
In \cref{fig:app-feature-weights}, we present analysis of feature weights from a \HTR\ experiment, showing how the feature weights change as querying continues, averaged over 20 trials. 
The purpose is to show that it is possible to analyze the learned feature weights to help understand why \NOLSTAR\ algorithms perform the way they do. We choose \HTR\ here, but in principle any parameterized model could be temporally analyzed in a similar fashion.

In the BA example (\cref{fig:app-features-BA}), the LostReward feature, which takes order effects of querying into account (see \cref{sec:exp}), appears as the most positively weighted feature. 
This makes some intuitive sense: hubs accumulate larger values of lost reward over time, since many nodes brought in by other queries are connected after they eventually get queried. 
This means the hubs are ``missing out'' on reward and since they are in fact the best nodes to probe, this is reflected in the weights of the features. This implies that in a BA network, we expect that the degree and lost reward features will be correlated. This correlation provides a potential explanation for why degree is, somewhat counterintuitively, one of the least important features when querying a BA network: the degree of hub nodes is being accounted for in the lost reward feature, so degree itself does not have a strong impact.

In the BTER example (\cref{fig:app-features-BTER}), the normalized size of the connected component that a node is in is the most highly weighted feature. 
As a reminder, BTER networks are generated by constructing many ER networks (``communities'') of different sizes, where the size of the networks follows a power law distribution, then connecting them to one another.
This means there are a small number of very large and relatively well connected ``communities'', which are the best place to query to bring in new nodes.  

\begin{figure}[h]
	\centering
	\begin{subfigure}{0.5\textwidth}
		\includegraphics[width=\linewidth]{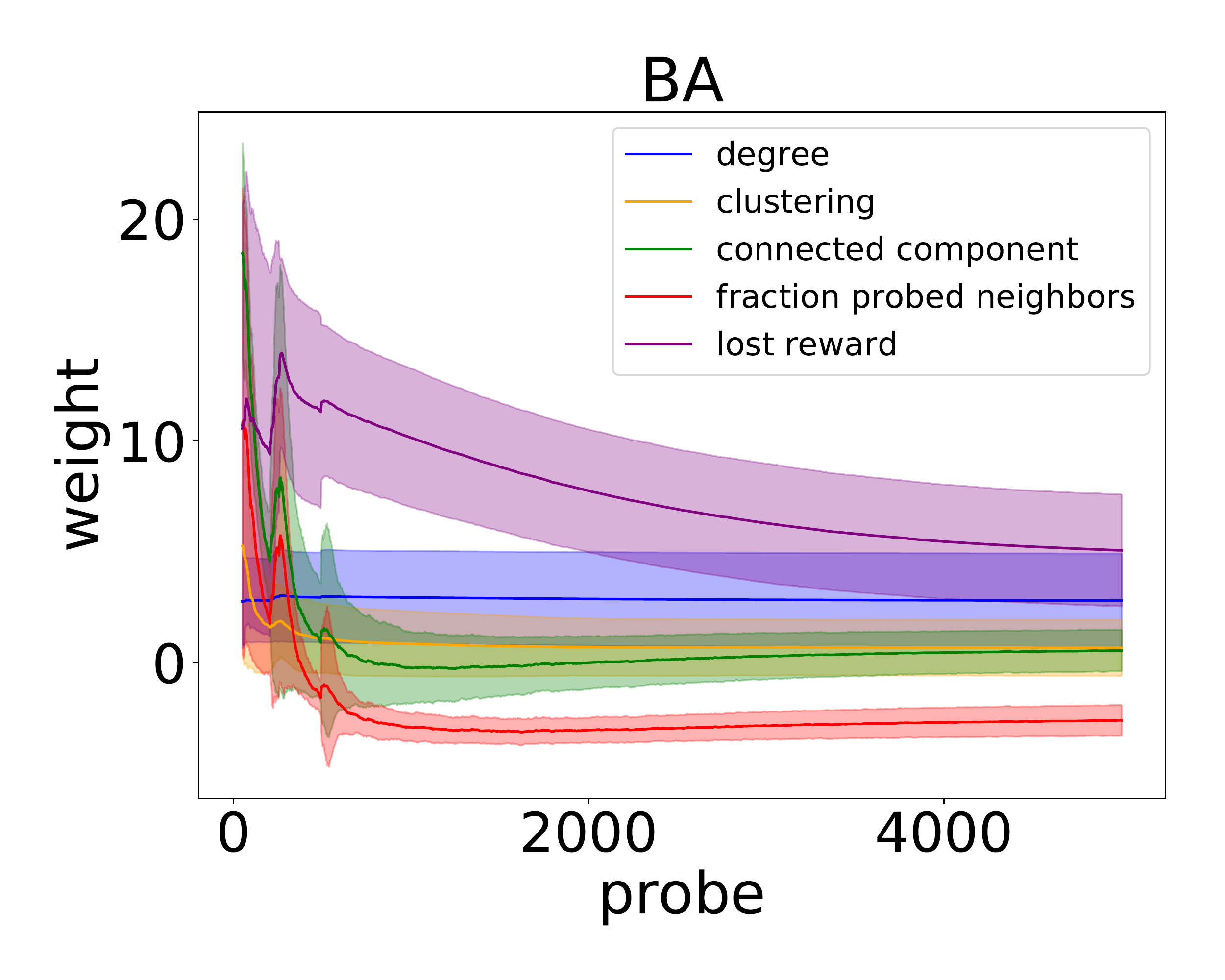}
		\caption{}
		\label{fig:app-features-BA}
	\end{subfigure}%
	\begin{subfigure}{0.5\textwidth}
		\includegraphics[width=\linewidth]{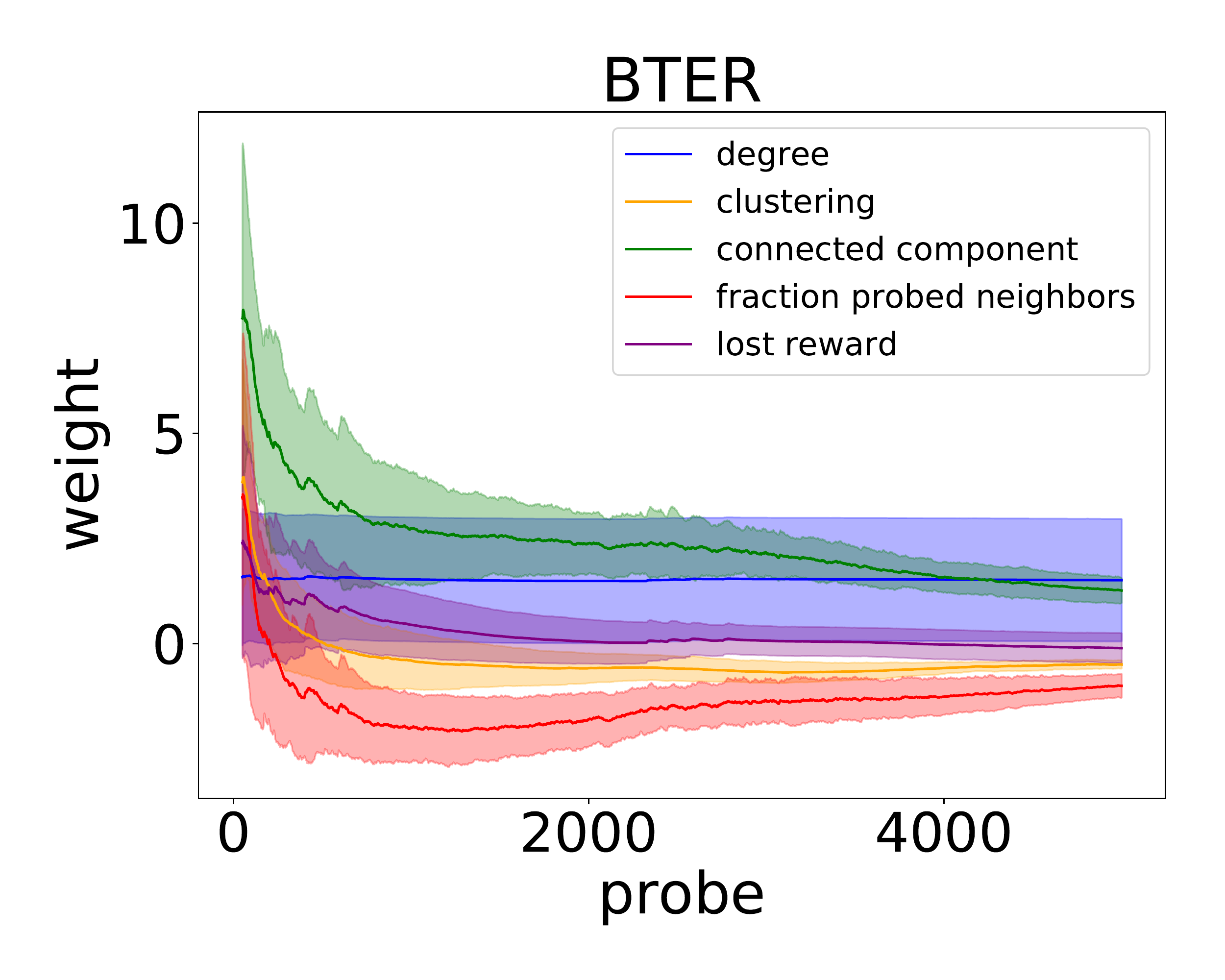}
		\caption{}
		\label{fig:app-features-BTER}
	\end{subfigure}
	\captionsetup{width=0.9\textwidth}
	\caption{Feature weights over time averaged over 20 runs of \HTR\ on (a) BA and (b) BTER networks.}
	\label{fig:app-feature-weights}
\end{figure}

\subsection{Random Walk Sampling}
In \cref{fig:app-cumulative-reward} we show results of \NOLSTAR\ algorithms starting from random walk samples. The sampling works as follows: a node is chosen uniformly at random, then a random walk from that node proceeds until the desired proportion of edges is discovered, jumping randomly 15\% of the time. The resulting performance is similar to what we saw in the node sampling with induction samples: \NOLSTAR\ algorithms are able to learn to match or outperform the heuristic methods in almost every case. However, the performance of \NOLOR\ appears to be more consistent on random walk samples, as evidenced by the fact that the standard deviation around the mean performance tends to be tighter. The performance of \HTR\ is approximately the same or even slightly worse (e.g. BTER) on the random walk samples. The low degree heuristic also appears to perform better on random walk samples.
\begin{figure*}[!ht]
\centering
\begin{subfigure}{0.33\textwidth}
\centering
	\includegraphics[width=\linewidth, height=3.5cm]{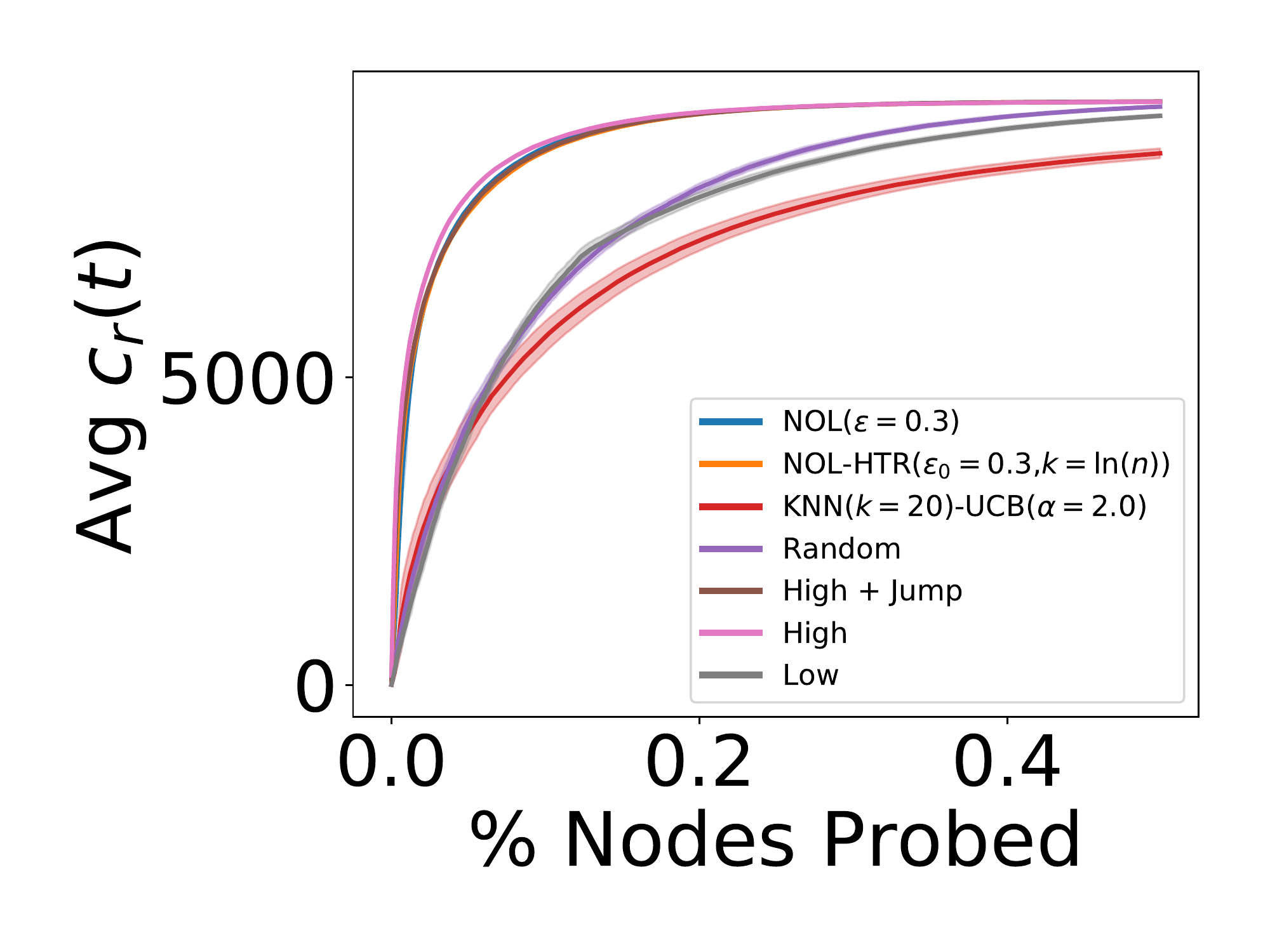}
    \vspace{-0.7cm}
    \caption{BA}
    \label{fig:cumulative-reward-walkjump-BA}
\end{subfigure}%
\begin{subfigure}{0.33\textwidth}
	\includegraphics[width=\linewidth, height=3.5cm]{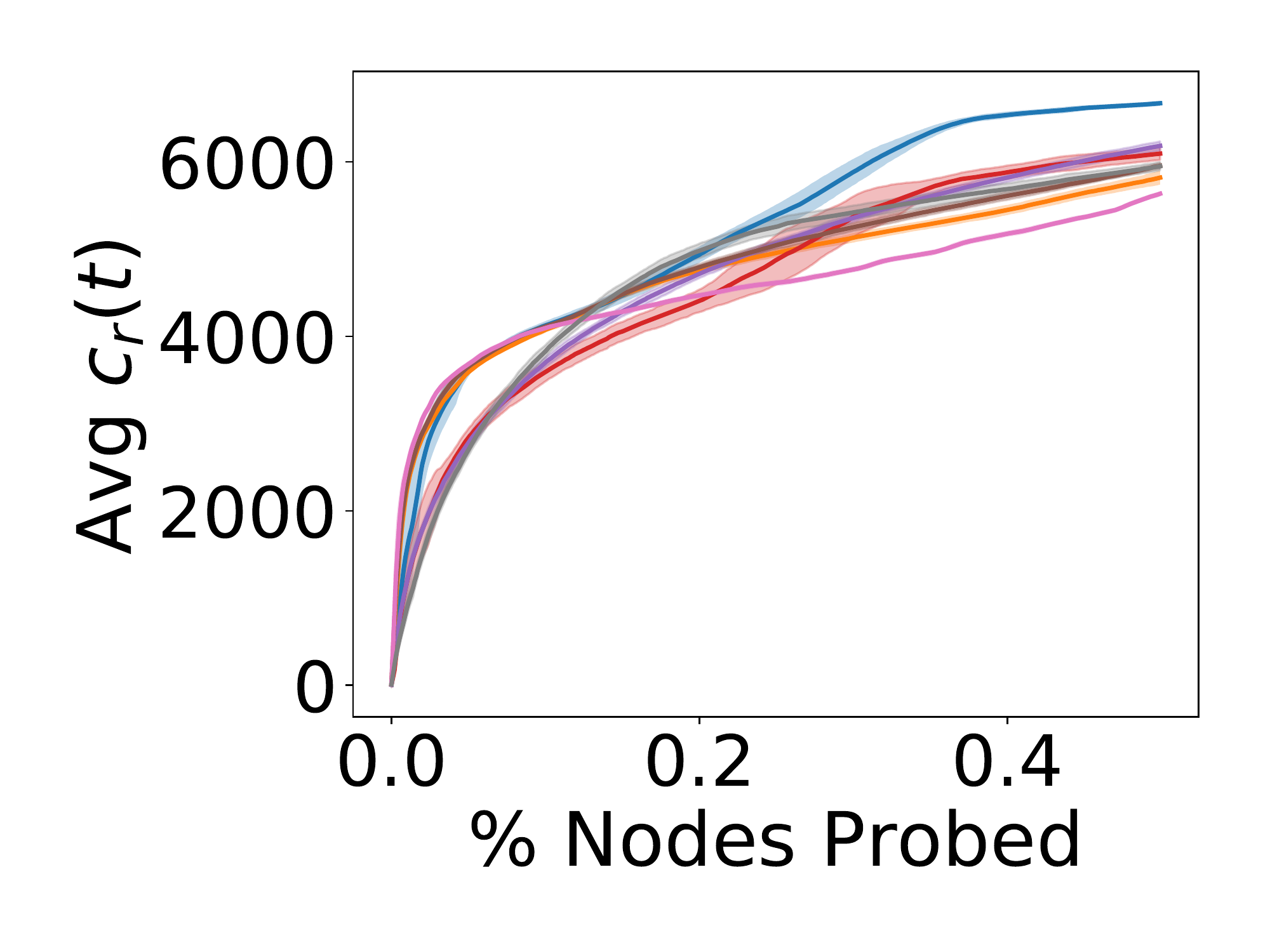}
    \vspace{-0.7cm}
    \caption{BTER}
    \label{fig:cumulative-reward-walkjump-BTER}
\end{subfigure}%
\begin{subfigure}{0.33\textwidth}
	\includegraphics[width=\linewidth, height=3.5cm]{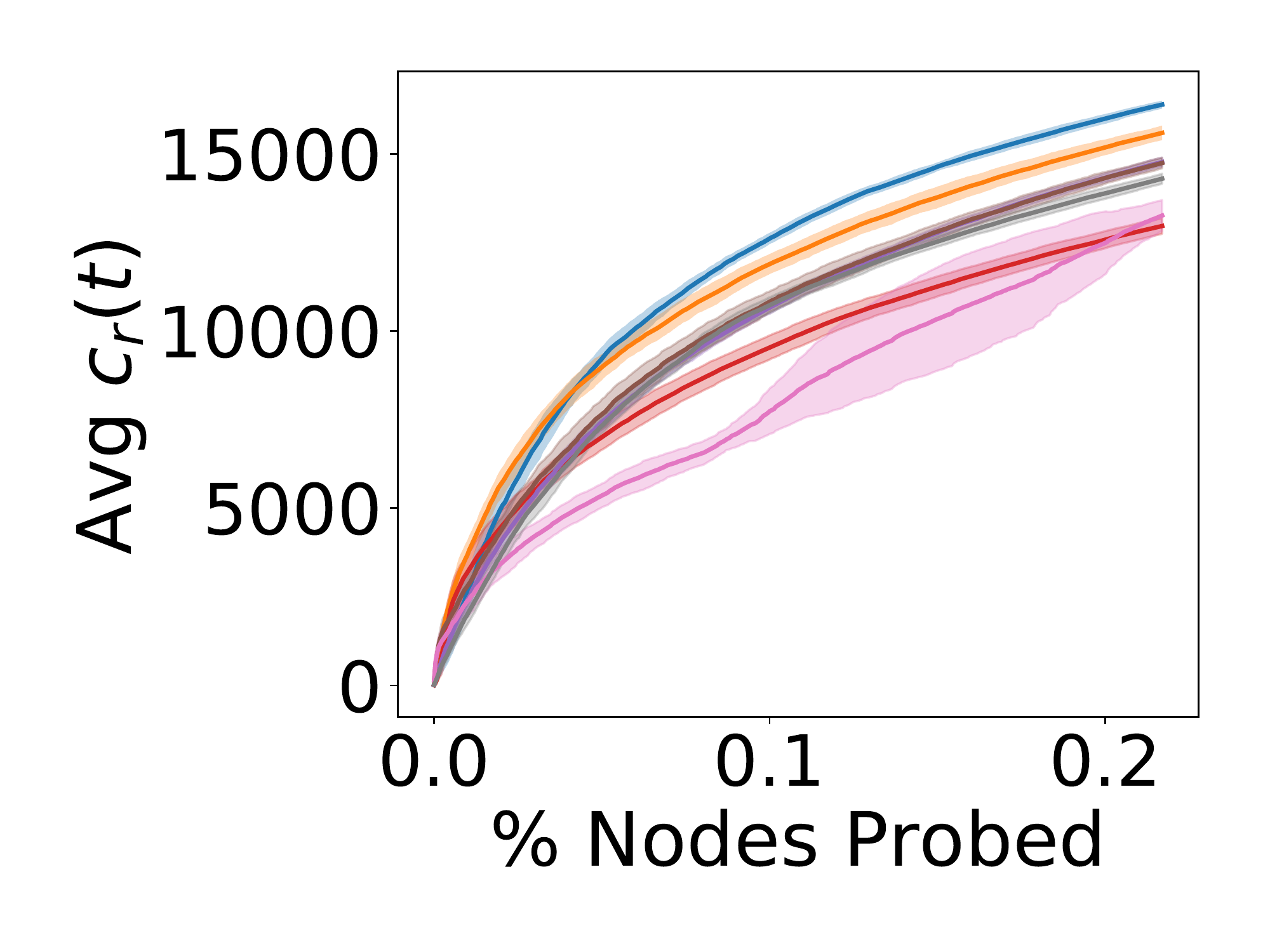}
    \vspace{-0.7cm}
    \caption{Cora}
    \label{fig:cumulative-reward-walkjump-cora}
\end{subfigure}
\begin{subfigure}{0.33\textwidth}
	\includegraphics[width=\linewidth, height=3.5cm]{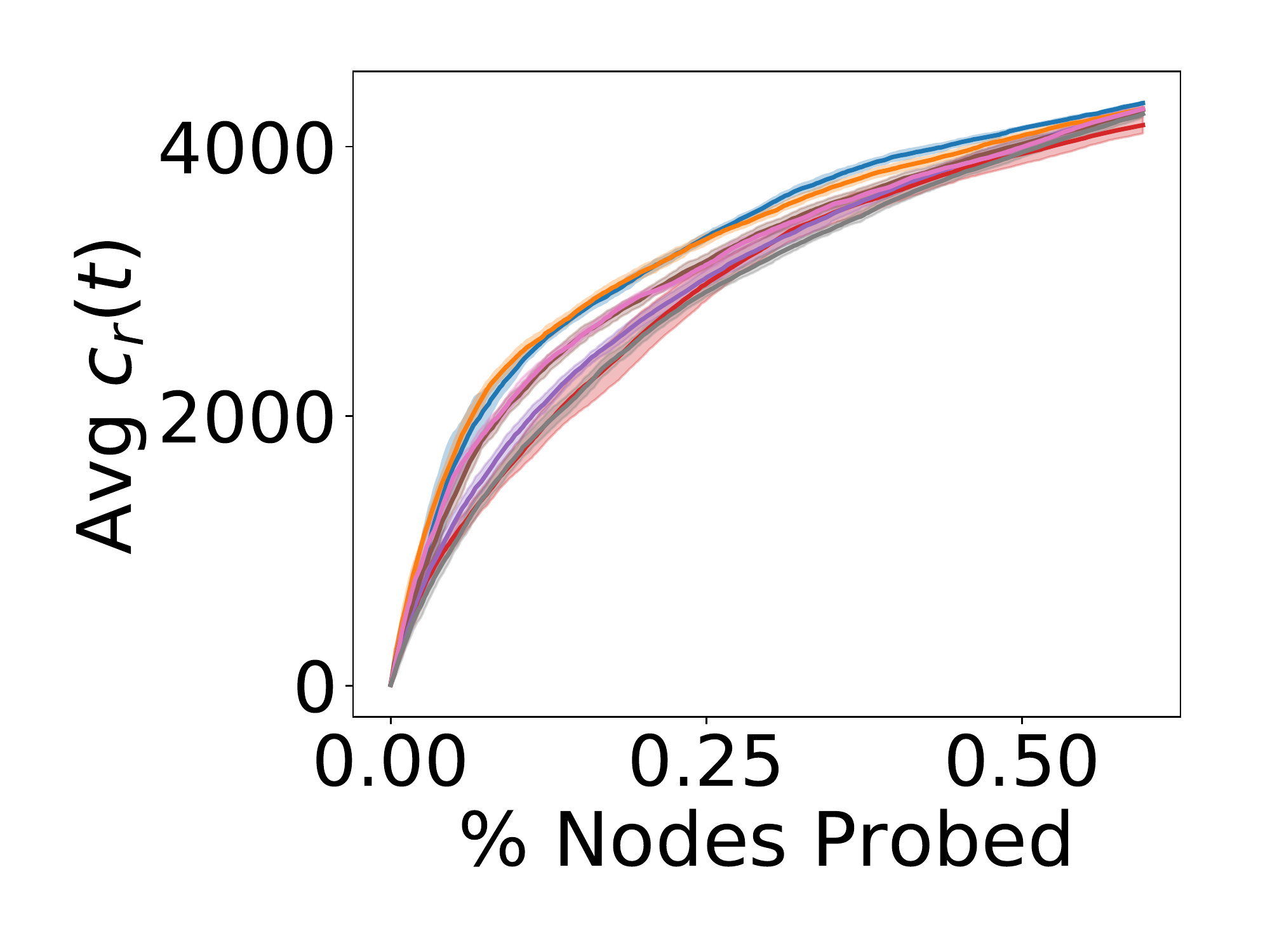}
    \vspace{-0.7cm}
    \caption{DBLP}
    \label{fig:cumulative-reward-walkjump-dblp}
\end{subfigure}%
\begin{subfigure}{0.33\textwidth}
	\includegraphics[width=\linewidth, height=3.5cm]{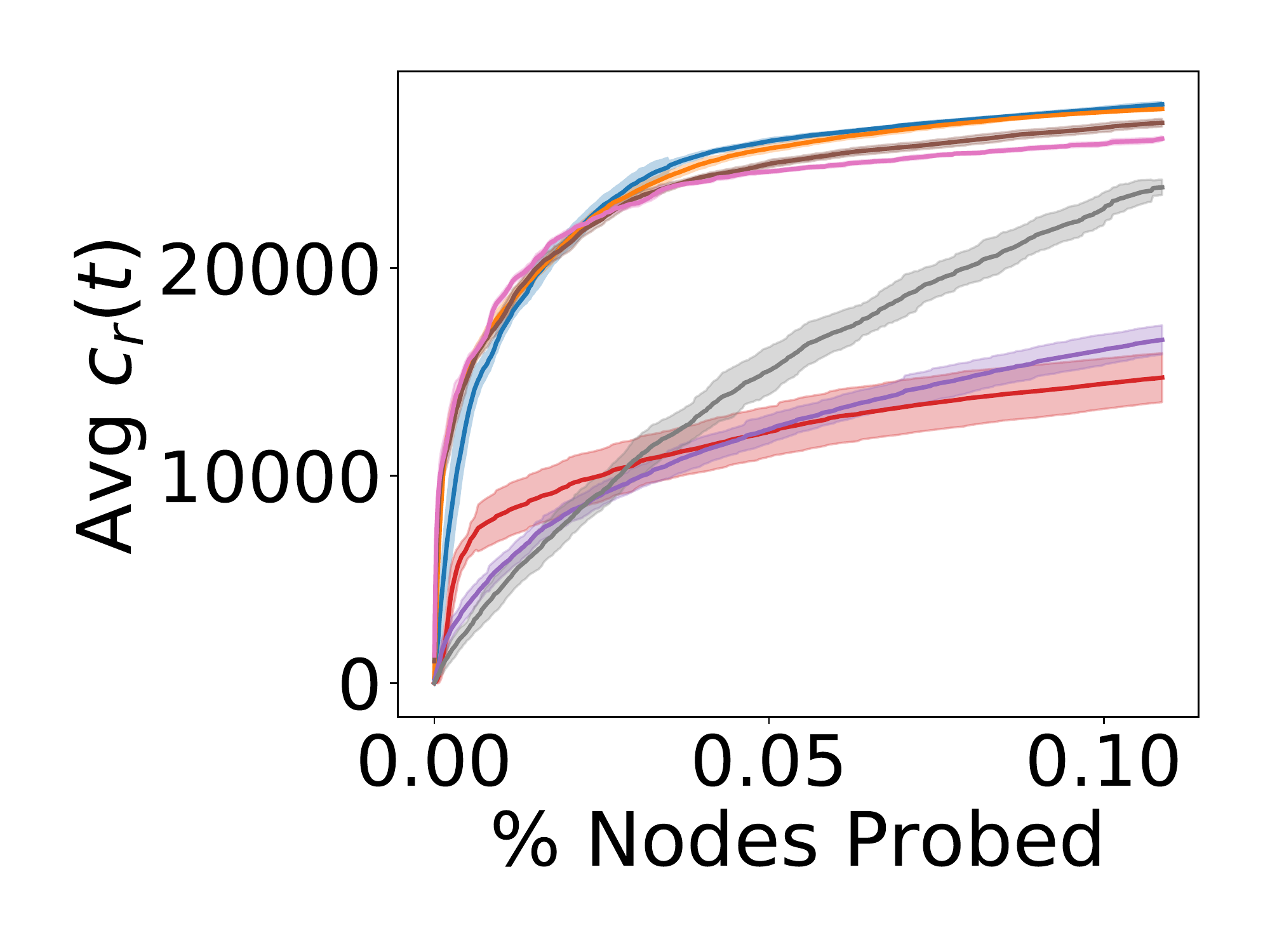}
    \vspace{-0.7cm}
    \caption{Enron}
    \label{fig:cumulative-reward-walkjump-enron}
\end{subfigure}%
\begin{subfigure}{0.33\textwidth}
	\includegraphics[width=\linewidth,height=3.5cm]{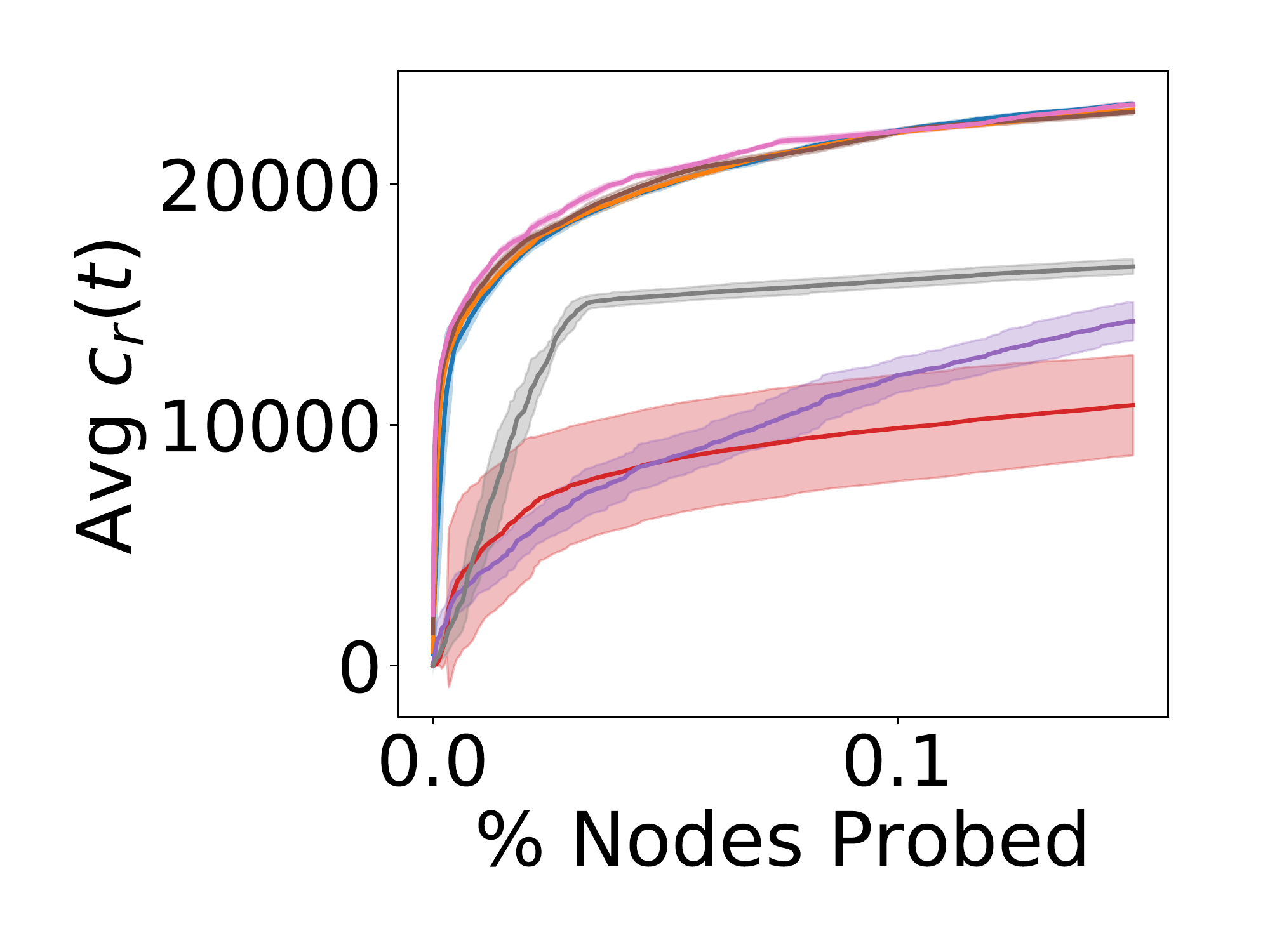}
    \vspace{-0.7cm}
    \caption{Caida}
    \label{fig:cumulative-reward-walkjump-caida}
\end{subfigure}
\vspace{-0.3cm}
    \captionsetup{width=0.9\textwidth}
    \caption{Results of running \NOLSTAR\ algorithms on the same networks as \cref{fig:cumulative-reward}, but starting from initial samples using random walk with jump sampling rather than node sampling with induction. Results are similar to those presented in the main text. In general, the performance of \NOLOR\ is more consistent on random walk samples, while the performance of \HTR\ is approximately the same or slightly worse (e.g. BTER). Interestingly, the low degree heuristic also appears to perform better on random walk samples.}
    \label{fig:app-cumulative-reward}
\end{figure*}

\subsection{Alternative Features}
In this section, we show some experiments using node2vec (\cite{node2vec}) as features, rather than our hand selected features.
Since node2vec is significantly more computationally expensive, we expect a tradeoff between computation time and performance increases due to more expressive features.
However, the results shown in \cref{fig:app-node2vec} present a mixed picture: In some cases, using the embedding features makes no improvement or can even degrade performance. Only in one case, the Cora network, do the node2vec features truly outperform the others, and even then only with one of the learning algorithms (\HTR). 
These results are inconclusive on the benefit of using embeddings as features and show that more experimentation and research needs to be done to understand the tradeoff between complexity of features and improvement in performance.

\begin{figure}[h]
	\centering
	\begin{subfigure}{0.35\textwidth}
		\includegraphics[width=\linewidth]{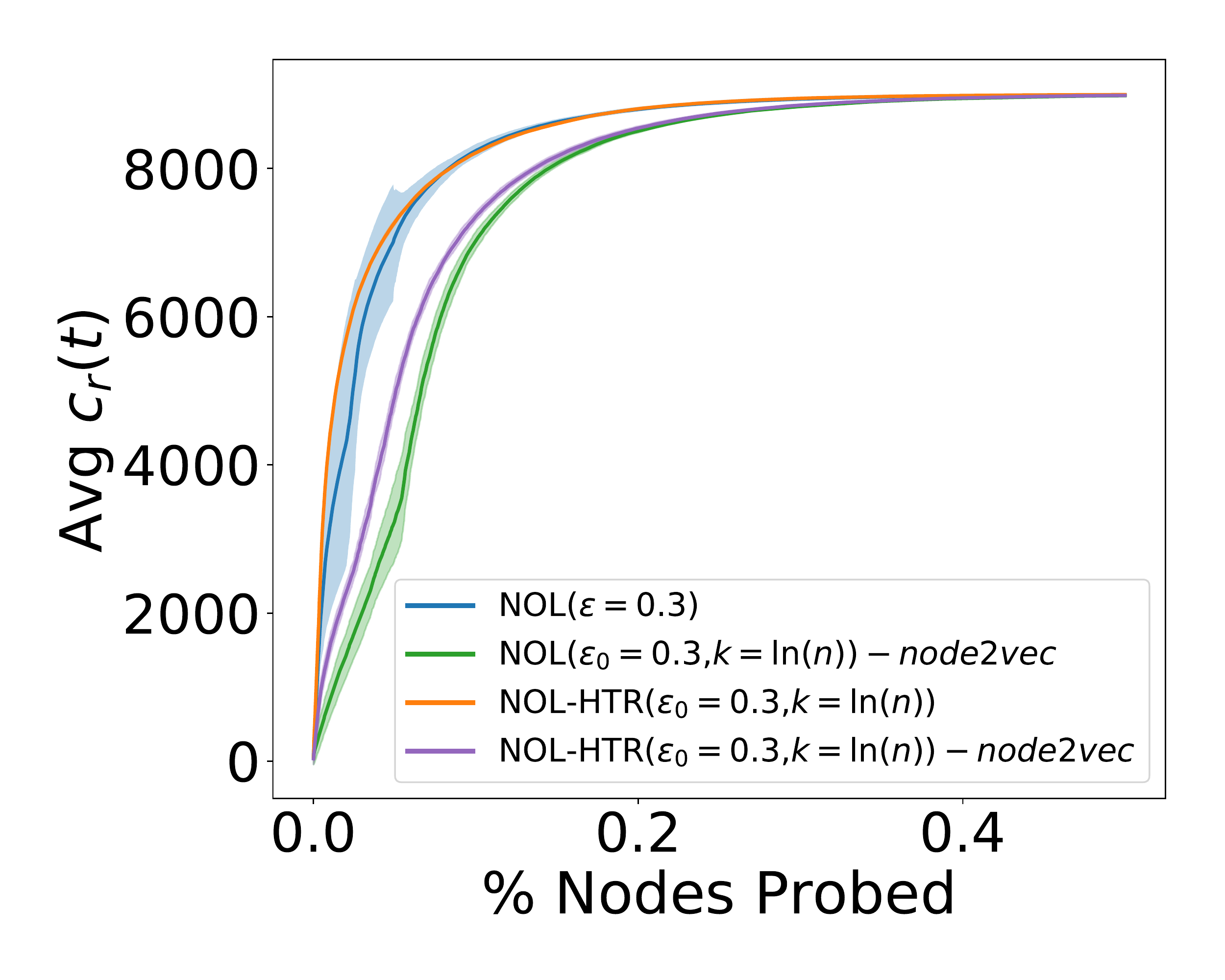}
		\caption{BA}
	\end{subfigure}%
	\begin{subfigure}{0.35\textwidth}
		\includegraphics[width=\linewidth]{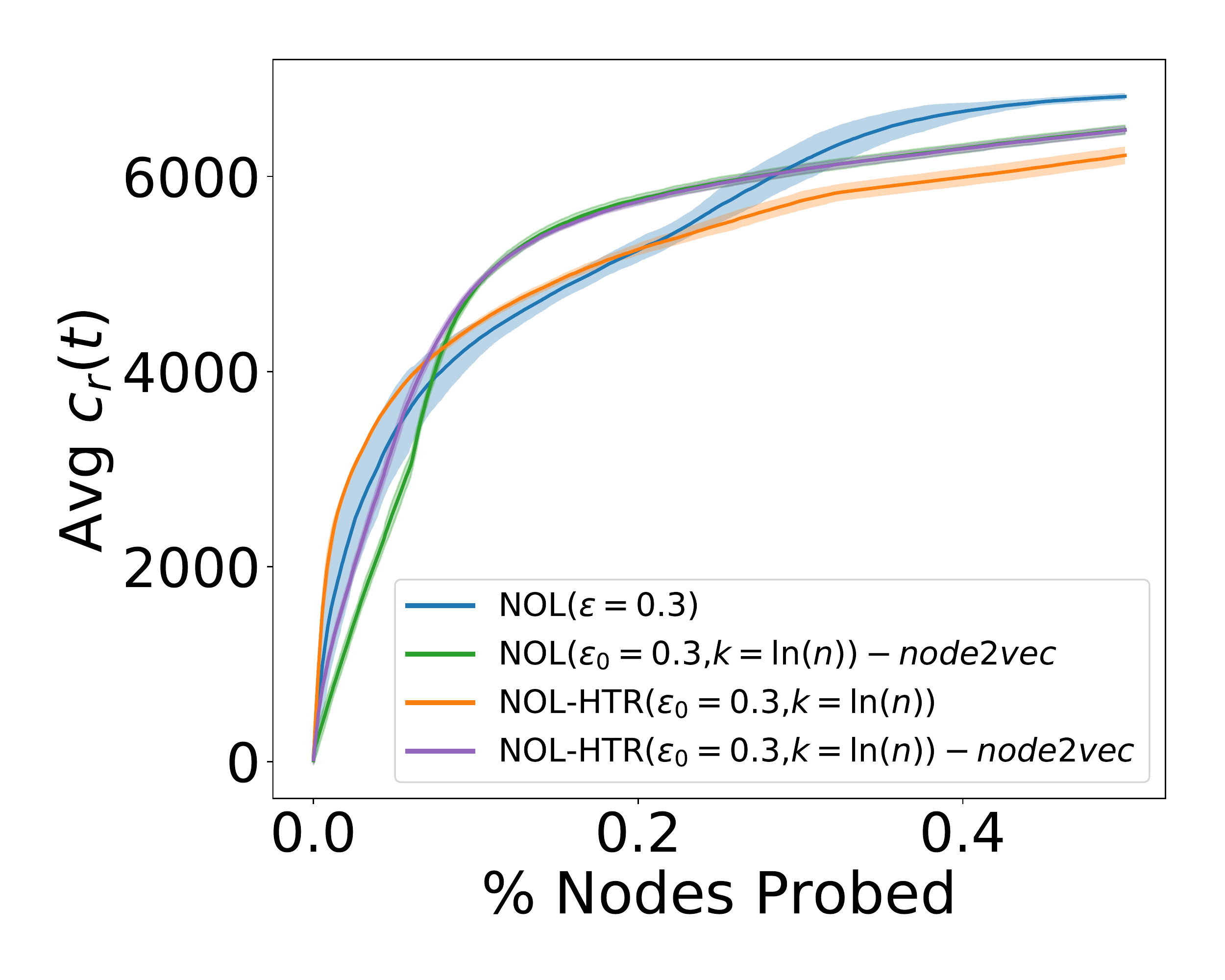}
		\caption{BTER}
	\end{subfigure}
	\begin{subfigure}{0.35\textwidth}
		\includegraphics[width=\linewidth]{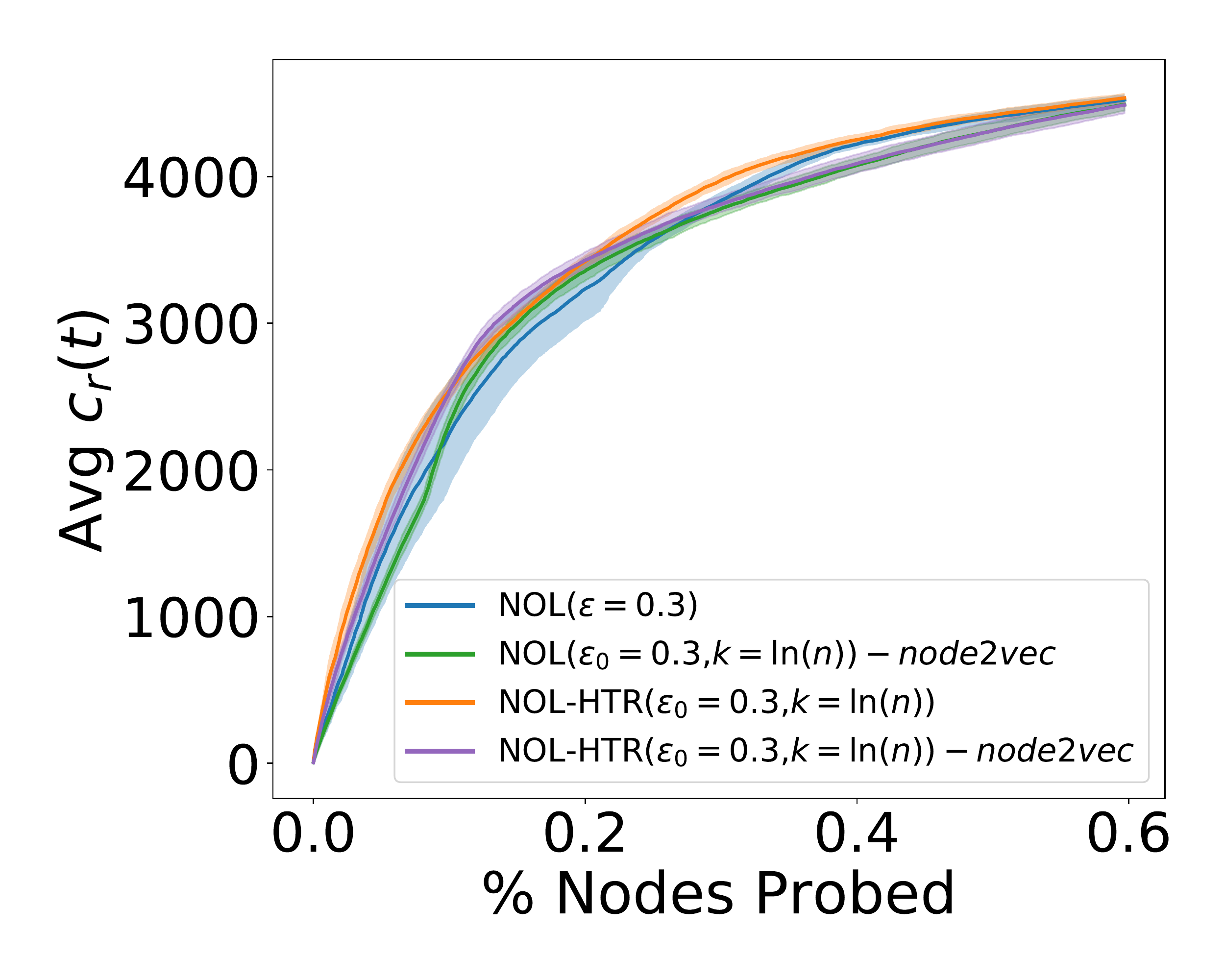}
		\caption{DBLP}
	\end{subfigure}%
	\begin{subfigure}{0.35\textwidth}
		\includegraphics[width=\linewidth]{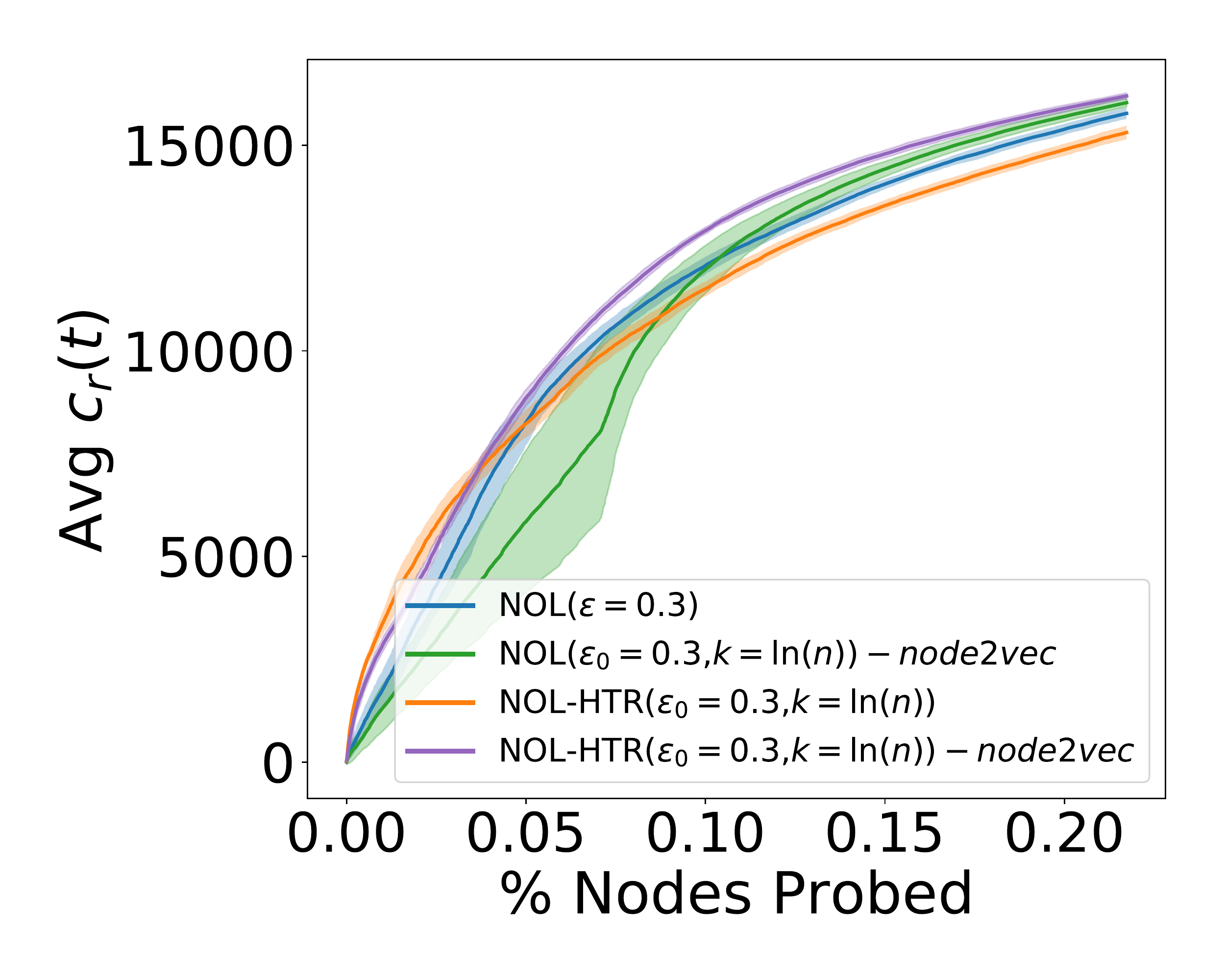}
		\caption{Cora}
	\end{subfigure}
	\captionsetup{width=0.9\textwidth}
	\caption{Cumulative reward results using node2vec features. Improvement using the embedding as features is not straightforward. In the BA and DBLP cases, performance is the same or worse. In the BTER case, performance improves towards the beginning, but does not decisively outperform the default features in the end. Only in Cora does it appear that \HTR\ with node2vec features decisively outperforms all of the other methods, but \NOLOR\ with node2vec only outperforms the other methods towards the end of the experiment.}
	\label{fig:app-node2vec}
\end{figure}

\clearpage

\end{backmatter}
\end{document}